\documentclass[10pt,journal,compsoc]{IEEEtran}
\usepackage[T1]{fontenc}
\usepackage[utf8]{inputenc}
\usepackage{array}
\usepackage{varioref}
\usepackage{booktabs}
\usepackage{enumitem}
\usepackage{multirow}
\usepackage{amsmath}
\usepackage{amsthm}
\usepackage{amssymb}
\usepackage{stmaryrd}
\usepackage{graphicx}
\usepackage{color}
\usepackage[algoruled,linesnumbered]{algorithm2e}
\usepackage[unicode=true,
 bookmarks=true,bookmarksnumbered=true,bookmarksopen=true,bookmarksopenlevel=1,
 breaklinks=false,pdfborder={0 0 0},pdfborderstyle={},backref=false,colorlinks=false]
 {hyperref}
\hypersetup{pdftitle={Skellam-SNMF},
 pdfauthor={Benoit Fuentes},
 pdfpagelayout=OneColumn, pdfnewwindow=true, pdfstartview=XYZ, plainpages=false}

\makeatletter

\providecommand{\tabularnewline}{\\}

 \let\oldforeign@language\foreign@language
 \DeclareRobustCommand{\foreign@language}[1]{%
   \lowercase{\oldforeign@language{#1}}}
\theoremstyle{plain}
\newtheorem{thm}{\protect\theoremname}
\theoremstyle{plain}
\newtheorem{prop}[thm]{\protect\propositionname}
\theoremstyle{plain}

\theoremstyle{remark}

\theoremstyle{definition}

\usepackage[caption=false,font=footnotesize]{subfig}
\usepackage{bbm}

\DeclareMathOperator*{\argmax}{arg\,max}
\DeclareMathOperator{\Pois}{Pois}

\DeclareMathOperator{\I}{I}
\DeclareMathOperator{\R}{R}

\DeclareMathOperator{\Skell}{Skell}

\DeclareMathOperator{\DiffNom}{DiffNomial}
\DeclareMathOperator{\Dir}{Dirichlet}
\DeclareMathOperator{\Gam}{Gamma}

\@ifundefined{showcaptionsetup}{}{%
 \PassOptionsToPackage{caption=false}{subfig}}
\usepackage{subfig}
\newcommand{\removelatexerror}{\let\@latex@error\@gobble}
\makeatother

\providecommand{\corollaryname}{Corollary}
\providecommand{\definitionname}{Definition}
\providecommand{\propositionname}{Proposition}
\providecommand{\remarkname}{Remark}
\providecommand{\theoremname}{Theorem}

\begin{document}

\title{Probabilistic semi-nonnegative matrix factorization: a Skellam-based
framework}

%,~\IEEEmembership{Senior~Member,~IEEE}
\author{Benoit~Fuentes and Gaël~Richard\IEEEcompsocitemizethanks{\IEEEcompsocthanksitem The authors are with LTCI, Institut Polytechnique de Paris, Télécom Paris,
Palaiseau, France, e-mail: \protect\href{mailto:bf@benoit-fuentes.fr}{bf@benoit-fuentes.fr},
\protect\href{mailto:gael.richard@telecom-paris.fr}{gael.richard@telecom-paris.fr}.}\thanks{This research leading to this paper has been conducted by Benoit Fuentes
while working in Smart Impulse, a french company.}}

%IEEE transactions on pattern analysis and machine intelligence
\markboth{Submitted for publication}{Benoit Fuentes \MakeLowercase{\emph{et al.}}: Probabilistic semi-nonnegative
matrix factorization: a Skellam-based framework}

\IEEEtitleabstractindextext{%
\begin{abstract}
We present a new probabilistic model to address semi-nonnegative matrix factorization (SNMF), called Skellam-SNMF. It is a hierarchical generative model consisting of prior components, Skellam-distributed hidden variables and observed data. Two inference algorithms are derived: Expectation-Maximization (EM) algorithm for maximum \emph{a posteriori} estimation and Variational Bayes EM (VBEM) for full Bayesian inference, including the estimation of parameters prior distribution. From this Skellam-based model, we also introduce a new divergence $\mathcal{D}$ between a real-valued target data $x$ and two nonnegative parameters $\lambda_{0}$ and $\lambda_{1}$ such that $\mathcal{D}\left(x\mid\lambda_{0},\lambda_{1}\right)=0\Leftrightarrow x=\lambda_{0}-\lambda_{1}$, which is a generalization of the Kullback-Leibler (KL) divergence. Finally, we conduct experimental studies on those new algorithms in order to understand their behavior and prove that they can outperform the classic SNMF approach on real data in a task of automatic clustering. 
\end{abstract}

}

\maketitle

\IEEEdisplaynontitleabstractindextext
\IEEEpeerreviewmaketitle

\IEEEraisesectionheading{\section{Introduction}\label{sec:introduction}}

\IEEEPARstart{M}{atrix} factorization, which consists in expressing or approximating a given matrix $\boldsymbol{X}$ as the product of two matrices $\boldsymbol{W}$ (called \emph{atoms} in this paper) and $\boldsymbol{\lambda}$ (called \emph{activations} in this paper), \emph{i.e.} $\boldsymbol{X}=\boldsymbol{W}\boldsymbol{\lambda}$ or $\boldsymbol{X}\approx\boldsymbol{W}\boldsymbol{\lambda}$, has been widely used in data analysis, signal processing and machine learning over many decades. There is a large number of techniques to address this problem, including principal component analysis (PCA), independent component analysis (ICA) \cite{Comon2010_book}, or Dictionary Learning \cite{Aharon2006_TSP} just to name a few. In some applications where observed matrix $\boldsymbol{X} \geq 0$, an additional constraint can be added to factors $\boldsymbol{W}$ and $\boldsymbol{\lambda}$ so they remain within the positive orthant, leading to the nonnegative matrix factorization problem (NMF) \cite{Lee2001_nips}. Beyond the innumerable applications of NMF that can be found in the literature, in fields such as astronomy \cite{Ren2017_TAJ}, audio signal processing \cite{Fevotte2009_NR}, bioinformatics \cite{Taslaman2012_PLOS}, text mining \cite{Huang2016_NIPS}, etc., there exists a great variety of theoretical work on NMF, focusing on different aspects of the problem \cite{Fu2019_SPM}. Without being exhaustive, one can mention studies on objective functions \cite{Fevotte2011_NR,Cichocki2006_LVA}, efficient algorithms \cite{Wang2010_SIAM} or probabilistic interpretation \cite{Cemgil2009_CIN,Hofman2001_ML,Shashanka2008_CIN}.

More recently, in domains such as energy efficiency \cite{Henriet2019_IEEE}, gene clustering \cite{Qi2009_BI}, template matching \cite{LeRoux2008_NIPS}, hidden representation learning \cite{Trigeorgis2014a_ICML}, or computational imaging \cite{Rousset2018_TCI}, a number of studies have made use of an alternative model called semi-nonnegative matrix factorization (SNMF), where observed matrix $\boldsymbol{X}$ is real-valued and where a nonnegativity constraint is added on the $\boldsymbol{\lambda}$ factor, leaving $\boldsymbol{W}$ unconstrained. SNMF was first introduced by Ding \emph{et. al. }\cite{Ding2010}. They define this problem as a classic optimization problem: given a matrix $\boldsymbol{X}\in\mathbb{R}^{I\times J}$, and a rank $K$, solve
\begin{equation}
\min_{\boldsymbol{W}\in\mathbb{R}^{I\times K},\boldsymbol{\lambda}\in\mathbb{R}^{K\times J}}\left\Vert \boldsymbol{X}-\boldsymbol{W}\boldsymbol{\lambda}\right\Vert _{F}^{2}\text{ such that }\boldsymbol{\lambda}\geq0,\label{eq:SNMF_ding_def}
\end{equation}
where $\left\Vert \cdot\right\Vert _{F}$ is the Frobenius norm. As for classical NMF \cite{Lee2001_nips}, this problem is solved by alternatively updating $\boldsymbol{W}$ and $\boldsymbol{\lambda}$. Update of $\boldsymbol{W}$ is performed via least square method and update of $\boldsymbol{\lambda}$ is performed via some multiplicative update which ensures the nonnegativity of $\boldsymbol{\lambda}$. Following this first article on SNMF, a few theoretical studies have been conducted in order to better understand this problem or to provide alternative solutions. In \cite{Chouh2015_LAIA} and \cite{Gillis2015_MAA}, the notion of semi-nonnegative rank of matrix $\boldsymbol{X}$ is introduced and SNMF algorithms under exact reconstruction constraint are developed. Gillis \emph{et. al.} \cite{Gillis2015_MAA} also put forward improvements to the original SNMF algorithm in order to overcome some former drawbacks such as numerical instability or slowness of convergence. Other studies focus on interpretability of parameters $\boldsymbol{W}$ and $\boldsymbol{\lambda}$ by adding extra regularization terms. In \cite{Dreisigmeyer2020_PRIA}, a constraint on $\boldsymbol{W}$ is introduced in order to minimize the maximum angle between any two columns of $\boldsymbol{W}$. In \cite{Henriet2019_IEEE}, the regularization term is designed to minimize the total variation of each row of $\boldsymbol{\lambda}$.

Although there seems to be a growing interest in this problem, knowledge about SNMF is limited compared to that of NMF. In order to make our contribution, in this paper we formulate SNMF as a statistical inference problem by developing a probabilistic framework suitable for this type of semi-nonnegative model. This framework, called Skellam-SNMF, is based on the Skellam distribution. It is a generalization to signed data of either Poisson NMF \cite{Cemgil2009_CIN} or probabilistic latent semantic analysis (PLSA) \cite{Hofman2001_ML} \textendash{} also known as probabilistic latent component analysis (PLCA) \cite{Shashanka2008_CIN} \textendash{} and its development into the fully-probabilistic latent Dirichlet allocation (LDA) model \cite{Blei2003_JMLR}. This will lead us to introduce a generalization to signed data of the Kullback-Leibler divergence, as an alternative to the classic Euclidean norm. We will also explain how to add priors on the model parameters as a way to perform regularization, and two inference algorithms will be developed in order to estimate factor matrices $\boldsymbol{W}$ and $\boldsymbol{\lambda}$: one for standard maximum \emph{a posteriori} estimation, and one for full Bayesian inference. Finally, we will see how to automatically infer the prior distribution of the parameters which shall open the path for online algorithms. By formulating it as a generalization of existing probabilistic models, the SNMF problem will benefit for future research from all improvements and enhancements that have been made on probalistic NMF. We think for instance of generalized tensor factorizations \cite{Yilmaz2011_NIPS}, dynamic models \cite{Mysore2012_ICML}, sophisticated \emph{ad hoc} models \cite{Fuentes2013_TASLP}, etc.

The paper is organized as follows. After having presented in section \ref{sec:Skellam-distribution} some properties about the Skellam distribution, the Skellam-SNMF model is introduced in section \ref{sec:Skellam-SNMF:-the-generative}. Sections \ref{sec:EM algorithm} and \ref{sec:VBEM-algorithm} are dedicated to the derivation of two inference algorithms. In section \ref{sec:Experimental-studies} we conduct experiments on toy examples in order to better understand the behavior of our algorithms and we compare Skellam-SNMF with other SNMF methods on real data on a simple clustering problem. Finally, we present our conclusions and ideas for future work in section \ref{sec:Conclusions}.

Before tackling the subject, let us present the notations that will be used in the sequel. The bold letters, whether upper of lower case, always refer to sets of scalars, including tensors or matrices. The letters $X$ and $Z$ are dedicated to observed data and hidden sources respectively. The letter $\lambda$ is always used for nonnegative parameters of Poisson or Skellam distributions. A bar an top indicates that the parameter is expressed as a function of other basic parameters (\emph{i.e.} $\bar{\lambda}_{s}=\sum_{n}\lambda_{sn}$). Finally, the letter $\boldsymbol{\theta}$ is used to designate a set of nonnegative parameters subject to normalization constraints.

\section{Skellam distribution\label{sec:Skellam-distribution}}

Skellam-SNMF is based on a linear source mixture model where individual
sources are modeled as Skellam random variables and we present in
this section important properties about this distribution. All proofs
are reported in the supplementary material. A Skellam random variable
(r.v.) $X$ is defined as the difference of two independent Poisson
random variables:
\begin{equation}
\begin{cases}
X_{0} & \sim\Pois\left(\lambda_{0}\right)\\
X_{1} & \sim\Pois\left(\lambda_{1}\right)
\end{cases}\Leftrightarrow X=X_{0}-X_{1}\sim\Skell\left(\lambda_{0},\lambda_{1}\right).
\end{equation}
Parameters $\lambda_{0}$ and $\lambda_{1}$ are nonnegative and mean
and variance of $X$ are given by
\begin{align}
\left\langle X\right\rangle  & =\lambda_{0}-\lambda_{1},\\
\text{Var}\left(X\right) & =\lambda_{0}+\lambda_{1}.
\end{align}
There exists several equivalent expressions for the Skellam distribution
\cite{Alzaid2010_MMSS} and the one that will be used in this paper
is the following:
\begin{equation}
P\left(X=x\right)=\frac{_{0}F_{1}\left(|x|+1,\lambda_{0}\lambda_{1}\right)}{\Gamma\left(|x|+1\right)}\prod_{s\in\left\{ 0,1\right\} }e^{-\lambda_{s}}\lambda_{s}^{\max\left(\left(-1\right)^{s}x,0\right)}\label{eq:Skellam-distribution}
\end{equation}
where $_{0}F_{1}$ is the confluent hypergeometric limit function
(which is closely related to the modified Bessel function of the first
kind) and where $x\in\mathbb{Z}$. It is easy to verify that this
distribution is simplified into the Poisson distribution if $\lambda_{1}=0$. 
A key property is that the sum of independant Skellam r.v. $Z_{n}\sim\Skell\left(\lambda_{0,n},\lambda_{1,n}\right)$
is a also a Skellam r.v.:
\begin{equation}
X=\sum_{n}Z_{n}\sim\Skell\left(\sum_{n}\lambda_{0,n},\sum_{n}\lambda_{1,n}\right).
\end{equation}
We refer $\left\{ Z_{n}\right\} $ as the \emph{hidden Skellam sources
}and $X$ as the \emph{observed mixture} or \emph{observed data}.
Besides, the underlying Poisson r.v. $\left\{ Z_{sn}\sim\Pois\left(\lambda_{sn}\right)\right\} _{s\in\left\{ 0,1\right\} ,n}$
such that $Z_{n}=Z_{0,n}-Z_{1,n}$ are called \emph{hidden Poisson
sources.}

Now, we are interested in the posterior distribution of those hidden
Poisson sources given the observed mixture, since it will be useful
during the derivation of the statistical inference algorithms used
later on. First, it can be proven that the expectation
of this posterior distribution is given by:
\begin{equation}
\begin{aligned}\left\langle Z_{sn}\mid X=x\right\rangle = & \lambda_{sn}\left[\frac{\max\left(\left(-1\right)^{s}x,0\right)}{\bar{\lambda}_{s}}+\right.\\
 & \left.\frac{\bar{\lambda}_{1-s}}{|x|+1+\sqrt{\bar{\lambda}_{0}\bar{\lambda}_{1}}\R_{|x|+1}\left(2\sqrt{\bar{\lambda}_{0}\bar{\lambda}_{1}}\right)}\right]
\end{aligned}
\label{eq:Posterior-Expectation-Skellam-1}
\end{equation}
where $\bar{\lambda}_{s}=\sum_{n}\lambda_{sn}$ for $s\in\left\{ 0,1\right\} $
and where $\textrm{R}_{x}\left(z\right)$ is the ratio of modified
Bessel functions of the first king \cite{Gautschi1978_MOC}:
\begin{equation}
\R_{x}\left(z\right)=\frac{\I_{x+1}\left(z\right)}{\I_{x}\left(z\right)}.\label{eq:bessel_ratio}
\end{equation}
Then, using Bayes rule, one can give the full posterior distribution
of $\boldsymbol{Z}=\left\{ Z_{sn}\right\} $ with respect to $X$
and parameters $\boldsymbol{\lambda}=\left\{ \lambda_{sn}\right\} $:
\begin{equation}
P\left(\boldsymbol{Z}=\boldsymbol{z}\mid X=x\right)=p\left(\boldsymbol{z};\boldsymbol{\lambda},x\right)
\end{equation}
with
\begin{equation}
p\left(\boldsymbol{z};\boldsymbol{\lambda},x\right)=\mathrm{D}\left(\boldsymbol{\lambda},x\right)\frac{\prod_{sn}\lambda_{sn}^{z_{sn}}}{\prod_{sn}z_{sn}!}\mathbbm1_{\left\{ x=\sum_{n}z_{0,n}-z_{1,n}\right\} }\label{eq:diffnomial_def_1}
\end{equation}
where $\mathbbm1$ is the indicator function and where 
\begin{equation}
\mathrm{D}\left(\boldsymbol{\lambda},x\right)=\frac{\prod_{s}\left(\sum_{n}\lambda_{sn}\right)^{-\max\left(\left(-1\right)^{s}x,0\right)}}{_{0}F_{1}\left(|x|+1,\prod_{s}\sum_{n}\lambda_{sn}\right)}\Gamma\left(|x|+1\right)\label{eq:diffnomial_def_2}
\end{equation}
is the normalization factor. To our knowledge, such a distribution
has not yet been introduced in the literature. We decide to name it
the \textbf{\emph{diffnomial}}\textbf{ }distribution, as a reference
to the equivalent \emph{multinomial }law in the Poisson mixture case\footnote{It is a well known result that if $Z_{n}\sim\Pois\left(\lambda_{n}\right)$
for $n=1\dots N$ and if $X=\sum_{n}Z_{n}$, then $\left\{ Z_{n}\right\} \mid X$
follows a multinomial distribution.} :
\begin{equation}
\left(\boldsymbol{Z}\mid X=x\right)\sim\DiffNom\left(x,\boldsymbol{\lambda}\right).
\end{equation}
It is easy to verify that the diffnomial law is indeed simplified
into a multinomial law if $\lambda_{1,n}$ and $z_{1,n}$ are set
to $0$ for all $n$.

Now we have presented all necessary background preliminaries, the
Skellam-SNMF model can be introduced.

\section{Skellam-SNMF: the generative model\label{sec:Skellam-SNMF:-the-generative}}

We aim at approximating a matrix $\boldsymbol{X}$ as a factorization of two matrices $\boldsymbol{X}\approx\boldsymbol{W}\boldsymbol{\lambda}$ where $\boldsymbol{W}\in\mathbb{R}^{I\times K}$ contains real values and $\boldsymbol{\lambda}\in\mathbb{R}_{+}^{K\times J}$ only nonnegative ones. The main idea in Skellam-NMF is to express atoms matrix $\boldsymbol{W}$ as the difference between two nonnegative matrices $\boldsymbol{W}=\boldsymbol{\theta}_{0}-\boldsymbol{\theta}_{1}$ and then to consider that each coefficient $X_{ij}$ is drawn from a Skellam distribution $X_{ij}\sim\Skell\left(\left[\boldsymbol{\theta}_{0}\boldsymbol{\lambda}\right]_{ij},\left[\boldsymbol{\theta}_{1}\boldsymbol{\lambda}\right]_{ij}\right)$. An appropriate estimator for parameters $\boldsymbol{\theta}_{0}$, $\boldsymbol{\theta}_{1}$ and $\boldsymbol{\lambda}$ will try to make the expected value of Skellam distribution $\left[\boldsymbol{\theta}_{0}\boldsymbol{\lambda}\right]_{ij}-\left[\boldsymbol{\theta}_{1}\boldsymbol{\lambda}\right]_{ij}$ as closed as possible to the observed data and then have the best possible approximation $\boldsymbol{X}\approx\hat{\boldsymbol{X}}=\boldsymbol{\theta}_{0}\boldsymbol{\lambda}-\boldsymbol{\theta}_{1}\boldsymbol{\lambda}=\boldsymbol{W}\boldsymbol{\lambda}$. With this generative model, only integers are allowed for the coefficients of $\boldsymbol{X}$. For real-valued data, the idea is to consider $\boldsymbol{X}$ as the mean of $M$ Skellam-distributed matrices $\boldsymbol{X}=\frac{1}{M}\sum_{m=1}^{M}\boldsymbol{X}^{m}$ and then make $M$ tends towards $\infty$.

\subsection{Normalization constraints on atoms}

From now on, we gather the two matrices $\boldsymbol{\theta}_{0}$ and $\boldsymbol{\theta}_{1}$ in a single tensor $\boldsymbol{\theta}=\left\{ \theta_{sik}\right\} _{s\in\left\{ 0,1\right\} ,i=1\dots I,k=1\dots K}$ also called \emph{atoms} herein. We decide to add the following normalization constraint on $\boldsymbol{\theta}$:

\begin{equation}
\forall k,\ \sum_{si}\theta_{sik}=1.\label{eq:atoms normalization constraint}
\end{equation}
In order to notify such a constraint, we will use notation $\theta_{si\mid k}$ instead of $\theta_{sik}$. This presents many advantages. First of all, it overcomes an homogeneity flaw that happens when the observed data $\boldsymbol{X}$ has a physical dimension, such as Watts (W), lumen (lm), etc.: since $\hat{\boldsymbol{X}}=\boldsymbol{W}\boldsymbol{\lambda}$ should have the same physical dimension, it makes more sense to have one normalized factor with no dimension whatsoever and one factor that carries the physical dimension, than two factors that would be expressed in square root of the dimension. Then, a practical advantage is that this constraint leads to the following simplification
\begin{equation}
\sum_{ijks}\theta_{si\mid k}\lambda_{kj}=\sum_{kj}\lambda_{kj}\label{eq: normalization simplification}
\end{equation}
which facilitates the derivation of both inference algorithms presented in sections \ref{sec:EM algorithm} and \ref{sec:VBEM-algorithm}. It also naturally prevents any estimation algorithm from numerical stability problems, with for instance atoms tending towards very small values and activations tending towards very high ones. Finally, it removes a well known identifiability problem, namely the scale invariance between columns of $\boldsymbol{W}$ and rows of $\boldsymbol{\lambda}$. Note that the choice to apply a normalization constraint on atoms $\boldsymbol{\theta}$ is arbitrary and we could have normalized activations instead.

\subsection{Priors on parameters}

We consider the possibility of adding priors on parameters as a way to both get rid of all identifiability problems that might remain and to add regularization terms in the objective function to be optimized. This can help to find more relevant estimates for the parameters, depending on the application. In order to stay in a easy-to-compute probabilistic framework, we suggest the use of conjugate priors for the Skellam likelihood function, which happens to be Gamma priors for non-normalized parameters $\lambda_{kj}$ and Dirichlet priors for normalized parameters $\theta_{si\mid k}$.

\subsection{The generative model\label{subsec:The-generative-model}}

Now, we can detail the full generative model of Skellam-SNMF. In order to consider both cases according to whether $\boldsymbol{X}$ is composed of integers or real numbers, we let the number $M$ undefined. Just be aware that the two values of interest are $M=1$ for integer data or $M\rightarrow\infty$ for real data. Note also that we are going to artificially over-parameterize our model by defining $M$ atoms tensors and $M$ activations matrices. This will be discussed at the end of this section. The first step of the generative model is to draw parameters from their prior distributions:
\begin{align}
\forall m=1\dots M,\ \left\{ \theta_{si\mid k}^{m}\right\} _{si} & \sim\Dir\left(\left\{ \alpha_{\varphi\left(s,i,k\right)}\right\} _{si}\right),\label{eq:gp-atoms-prior}\\
\lambda_{kj}^{m} & \sim\Gam\left(\alpha_{\upsilon\left(k,j\right)},\beta_{\omega\left(k,j\right)}\right).\label{eq:gp-activations-prior}
\end{align}
Here, $\boldsymbol{\alpha}=\left\{ \alpha_{a}\right\} $ and $\boldsymbol{\beta}=\left\{ \beta_{b}\right\} $ are two sets of non-negative shape and rate hyperparameters, and $\varphi$, $\upsilon$ and $\omega$ are functions that map the parameters to the hyperparameters. Using such maps allows us to keep the possibility for several parameters to share a same hyperparameter, reducing then their number. Later, we will see how hyperparameters can be learned from the data, and such feature can be useful in order to avoid overfitting. However, we do not permit Gamma and Dirichlet parameters to share a same shape hyperparameter and one must have
\begin{equation}
\left\{ \varphi\left(s,i,k\right)\right\} _{s,i,k}\cap\left\{ \upsilon\left(k,j\right)\right\} _{k,l}=\emptyset.
\end{equation}

The second step is to draw Poisson hidden variables (or hidden sources) depending on the parameters:
\begin{equation}
Z_{sikj}^{m}\sim\Pois\left(\bar{\lambda}_{sikj}^{m}\right)
\end{equation}
with
\begin{equation}
\bar{\lambda}_{sikj}^{m}=\theta_{si\mid k}^{m}\lambda_{kj}^{m}.\label{eq:SNMF-poisson-source-param-model}
\end{equation}

Finally, observed data are computed as:
\begin{equation}
X_{ij}=\frac{1}{M}\sum_{mk}Z_{s=0,ikj}^{m}-\sum_{mk}Z_{s=1,ikj}^{m},
\end{equation}
leading to Skellam independent random variables for $M\times$ the observed data:
\begin{equation}
MX_{ij}\sim\Skell\left(\sum_{m}\bar{\lambda}_{s=0,ij}^{m},\sum_{m}\bar{\lambda}_{s=1,ij}^{m}\right)\label{eq:gp-MX-skell-def}
\end{equation}
with
\begin{equation}
\bar{\lambda}_{sij}^{m}=\sum_{k}\bar{\lambda}_{sikj}^{m}.\label{eq:SNMF-skellam parameters model}
\end{equation}
Possibly, we can also consider that some data are missing, meaning that $\left\{ X_{ij}\right\} $ is observed only for a subset 
\begin{equation}
\mathcal{O}\subset\left\{ 1,\dots I\right\} \times\left\{ 1,\dots J\right\} 
\end{equation}
 of indexes $\left(i,j\right)$. We redefine then the set of observed data as
\begin{equation}
\boldsymbol{X}_{\mathcal{O}}=\left\{ X_{ij}\right\} _{ij\in\mathcal{O}}.
\end{equation}

The reason for over-parameterizing the model by drawing $M$ independent pairs of atoms and activations is that otherwise the log-prior probability of the parameters would become negligible compared to the log-likelihood of the data as $M$ grows, making the addition of priors useless. With $M$ draws, priors are ``counted'' $M$ times, which solves the problem. Now, the trick to getting back to a single atoms factor and a single activations factor is to constrain, when we seek to infer the parameter values, the estimates of $\boldsymbol{\theta}^{m}$ and $\boldsymbol{\lambda}^{m}$ to be pairwise equals,\emph{ i.e. }$\exists\left(\boldsymbol{\theta},\boldsymbol{\lambda}\right),\forall m,\ \left(\boldsymbol{\theta}^{m},\boldsymbol{\lambda}^{m}\right)=\left(\boldsymbol{\theta},\boldsymbol{\lambda}\right)$.

At this point, any statistical inference algorithms can be applied in order to estimate the best value for the parameters given observed data $\boldsymbol{X}$. In the two following sections, we focus on the Expectation-Maximization (EM) algorithm and the Variational Bayes EM algorithm.

\section{Skellam-SNMF with EM algorithm\label{sec:EM algorithm}}

\subsection{Objective function and emergence of a new divergence}

In order to estimate the parameters of our model, one can use the Expectation-Maximization algorithm \cite{Dempster1977_JRSS} which aims at finding a local maximum of the log-posterior probability of the parameters given the data. In the case of the generative model presented in previous section, the Bayes rule can be used to compute it:
\begin{equation}
\begin{aligned}\ln P & \left(\left\{ \boldsymbol{\theta}^{m}\right\} ,\left\{ \boldsymbol{\lambda}^{m}\right\} \mid M\boldsymbol{X}\right)=\ln P\left(M\boldsymbol{X}\mid\left\{ \boldsymbol{\theta}^{m}\right\} _{m},\left\{ \boldsymbol{\lambda}^{m}\right\} _{m}\right)\\
 & +\sum_{m}\ln P\left(\boldsymbol{\theta}^{m}\right)+\sum_{m}\ln P\left(\boldsymbol{\lambda}^{m}\right)+cst
\end{aligned}
\end{equation}
where $cst$ does not depend on the parameters. Each of the other terms can be computed using equations \eqref{eq:gp-atoms-prior}, \eqref{eq:gp-activations-prior}, \eqref{eq:gp-MX-skell-def} and the definition of the corresponding distributions. If now we add a normalization factor $\frac{1}{M}$ so that this quantity does not tend toward $-\infty$ as $M$ goes to $+\infty$, and if we consider, as justified before, only values of parameters such as $\forall m,\left(\boldsymbol{\theta}^{m},\boldsymbol{\lambda}^{m}\right)=\left(\boldsymbol{\theta},\boldsymbol{\lambda}\right)$, we can define the objective function that the EM algorithm will optimize as:
\begin{align}
f_{\boldsymbol{X}}^{M}\left(\boldsymbol{\theta},\boldsymbol{\lambda}\right)= & \frac{1}{M}\ln P\left(\left\{ \boldsymbol{\theta}^{m}=\boldsymbol{\theta}\right\} _{m},\left\{ \boldsymbol{\lambda}^{m}=\boldsymbol{\lambda}\right\} _{m}\mid M\boldsymbol{X}\right)\nonumber \\
= & \mathcal{L}_{\boldsymbol{X}}^{M}\left(\boldsymbol{\theta},\boldsymbol{\lambda}\right)\nonumber \\
 & +\sum_{kj}\left(\alpha_{\upsilon\left(k,j\right)}-1\right)\ln\lambda_{kj}-\beta_{\omega\left(k,j\right)}\lambda_{kj}\nonumber \\
 & +\sum_{sik}\left(\alpha_{\varphi\left(s,i,k\right)}-1\right)\ln\theta_{si\mid k} + cst/M\label{eq:EM-full-objective-function}
\end{align}
where 
\begin{equation}
\mathcal{L}_{\boldsymbol{X}}^{M}\left(\boldsymbol{\theta},\boldsymbol{\lambda}\right)=\frac{1}{M}\ln P\left(M\boldsymbol{X}\mid\left\{ \boldsymbol{\theta}^{m}=\boldsymbol{\theta}\right\} _{m},\left\{ \boldsymbol{\lambda}^{m}=\boldsymbol{\lambda}\right\} _{m}\right)\label{eq:EM-LM-def}
\end{equation}
 can be interpreted as a data fitting term, and the other terms as regularization terms ($cst/M$ is ignored thereafter). If $\boldsymbol{X}$ contains integer values ($M=1$), equation \eqref{eq:Skellam-distribution} gives:
\begin{equation}
\begin{aligned}\mathcal{L}_{\boldsymbol{X}}^{1}\left(\boldsymbol{\theta},\boldsymbol{\lambda}\right)= & \sum_{ij\in\mathcal{O}}\ln\frac{_{0}F_{1}\left(\left|X_{ij}\right|+1,\sigma_{ij}\right)}{\Gamma\left(\left|X_{ij}\right|+1\right)}+\\
 & \sum_{s}-\bar{\lambda}_{sij}+\max\left(\left(-1\right)^{s}X_{ij},0\right)\ln\bar{\lambda}_{sij}
\end{aligned}
\end{equation}
with
\begin{equation}
\sigma_{ij}=\bar{\lambda}_{s=0,ij}\bar{\lambda}_{s=1,ij}.\label{eq:SNMF-pi_def}
\end{equation}
and $\bar{\lambda}_{sij}$ given by equations \eqref{eq:SNMF-poisson-source-param-model} and \eqref{eq:SNMF-skellam parameters model}. If $\boldsymbol{X}$ is real-valued ($M=+\infty$), it can be proven using asymptotic expansion of $_{0}F_{1}$ (see supplementary material) that:
\begin{equation}
\mathcal{L}_{\boldsymbol{X}}^{\infty}\left(\boldsymbol{\theta},\boldsymbol{\lambda}\right)=-\sum_{ij\in\mathcal{O}}\mathcal{D}\left(X_{ij}\mid\bar{\lambda}_{s=0,ij},\bar{\lambda}_{s=1,ij}\right)\label{eq:EM-log-likelihood-limit}
\end{equation}
with
\begin{equation}
\begin{aligned}\mathcal{D} & \left(x\mid \lambda_{0},\lambda_{1}\right)=\sum_{s\in\left\{ 0,1\right\} }\lambda_{s}-\max\left(\left(-1\right)^{s}x,0\right)\ln \lambda_{s}\\
 & -\sqrt{x^{2}+4\lambda_{0}\lambda_{1}}+\left|x\right|\ln\left(\frac{\left|x\right|+\sqrt{x^{2}+4\lambda_{0}\lambda_{1}}}{2}\right).
\end{aligned}
\label{eq:EM-divergence-def}
\end{equation}
The function $\mathcal{D}\left(x\mid \lambda_{0},\lambda_{1}\right)$ can be seen as a divergence function: it is indeed always positive or null by construction and vanishes if and only if $x=\lambda_{0}-\lambda_{1}$. It is actually a generalization to signed data of the Kullback-Leibler (KL) divergence $\mathcal{D}_{\text{KL}}$ \cite{Lee2001_nips} since $\mathcal{D}\left(x\mid \lambda_{0},0\right)=\mathcal{D}_{\text{KL}}\left(x\mid \lambda_{0}\right)$ for nonnegative values of $x$. Note also that as for the KL divergence, it respects the following property:
\begin{equation}
\forall\mu>0,\ \mathcal{D}\left(\mu x\mid\mu \lambda_{0},\mu \lambda_{1}\right)=\mu\mathcal{D}\left(x\mid \lambda_{0},\lambda_{1}\right).
\end{equation}
To our knowledge, this divergence has never been introduced in the literature.

\subsection{Derivation of the EM algorithm\label{subsec:Derivation-of-EM}}

Each iteration of the EM algorithm consists of two steps. First the expectation step, were the log-likelihood of the complete data $\boldsymbol{Y}$ (observed and latent variables) is computed as well as its conditional expectation given current estimates for the parameters. Then the maximization step, where this last quantity is maximized with respect to the parameters. For the definition of $\boldsymbol{Y}$, we can either include or exclude missing data and latent sources that are linked to them. We decide to include them for a practical reason:  when deriving the algorithm, it allows to perform simplification \eqref{eq: normalization simplification} at some point, and without it, we would not have a simple closed form solution in the maximization stage. The downside in return is that it might slow down the speed of convergence, since it is a known feature of the EM algorithm that the more hidden variables compared to number of observed data, the slowest the convergence.
Curious readers may refer to the supplementary material, where the derivation of the EM algorithm is fully detailed. The computation is quite straightforward once the formula of the posterior expectation of the hidden sources \eqref{eq:Posterior-Expectation-Skellam-1} is known. The resulting update rules for the parameters are summarized in Algorithm \ref{EM-algorithm}.

\begin{figure*}[htb]
\centering
\begin{minipage}{.8\linewidth}
\begingroup
\removelatexerror% Nullify \@latex@error
\begin{algorithm}[H]
\BlankLine 

\KwIn{$\boldsymbol{X}$, $\mathcal{O},$ $M$, $\epsilon=0$ if shape
hyperparameters $\boldsymbol{\alpha}\geq1$ else $\epsilon>0$}

\BlankLine 

\KwOut{$\boldsymbol{\hat{\theta}}$ and $\boldsymbol{\hat{\lambda}}$}

\BlankLine 

initialize $\boldsymbol{\hat{\theta}}$ and $\boldsymbol{\hat{\lambda}}$ 

\Repeat{convergence}{

\tcc{Compute the model and the multiplicative updates}

$\bar{\lambda}_{sij}\leftarrow\sum_{k}\hat{\theta}_{si\mid k}\hat{\lambda}_{kj}$

optional: compute objective function $f_{\boldsymbol{X}}^{M}\left(\hat{\boldsymbol{\theta}},\hat{\boldsymbol{\lambda}}\right)$
(equation \eqref{eq:EM-full-objective-function})

$\sigma_{ij}\leftarrow\bar{\lambda}_{s=0,ij}\bar{\lambda}_{s=1,ij}$

$U_{sij}\leftarrow\begin{cases}
1 & \text{ if }ij\notin\mathcal{O}\\
\frac{\max\left(\left(-1\right)^{s}X_{ij},0\right)}{\bar{\lambda}_{sij}}+\frac{\bar{\lambda}_{1-s,ij}}{|X_{ij}|+1+\sqrt{\sigma_{ij}}\R_{|X_{ij}|+1}\left(2\sqrt{\sigma_{ij}}\right)} & \text{ if }ij\in\mathcal{O}\text{ and }M=1\\
\frac{\max\left(\left(-1\right)^{s}X_{ij},0\right)}{\bar{\lambda}_{sij}}+\frac{2\bar{\lambda}_{1-s,ij}}{|X_{ij}|+\sqrt{X_{ij}^{2}+4\sigma_{ij}}} & \text{ if }ij\in\mathcal{O}\text{ and }M=\infty
\end{cases}$

\begin{minipage}[t]{6cm}%
$U_{kj}^{\text{act}}\leftarrow\sum_{si}U_{sij}\hat{\theta}_{si\mid k}$%
\end{minipage}%
\begin{minipage}[t]{6cm}%
$U_{sik}^{\text{atoms}}\leftarrow\sum_{j}U_{sij}\hat{\lambda}_{kj}$%
\end{minipage}\BlankLine 

\tcc{Update parameters}

\begin{minipage}[t]{6cm}%
$\hat{\lambda}_{kj}\leftarrow\hat{\lambda}_{kj}U_{kj}^{\text{act}}+\alpha_{\upsilon\left(k,j\right)}-1$%
\end{minipage}%
\begin{minipage}[t]{6cm}%
$\hat{\theta}_{sik}\leftarrow\hat{\theta}_{si\mid k}U_{sik}^{\text{atoms}}+\alpha_{\varphi\left(s,i,k\right)}-1$%
\end{minipage}

\begin{minipage}[t]{6cm}%
$\hat{\lambda}_{kj}\leftarrow\min\left(\hat{\lambda}_{kj},\epsilon\right)$%
\end{minipage}%
\begin{minipage}[t]{6cm}%
$\hat{\theta}_{sik}\leftarrow\min\left(\hat{\theta}_{sik},\epsilon\right)$%
\end{minipage}

\begin{minipage}[t]{6cm}%
$\hat{\lambda}_{kj}\leftarrow\hat{\lambda}_{kj}/\left(1+\beta_{\omega\left(k,j\right)}\right)$%
\end{minipage}%
\begin{minipage}[t]{6cm}%
$\hat{\theta}_{si\mid k}\leftarrow\hat{\theta}_{si\mid k}/\sum_{s^{\prime}i^{\prime}}\hat{\theta}_{s^{\prime}i^{\prime}\mid k}$%
\end{minipage}

\BlankLine 

}

\caption{EM algorithm for Skellam-SNMF.\label{EM-algorithm}}
\end{algorithm}
\endgroup
\end{minipage}
\end{figure*}

\section{Full Bayesian inference\label{sec:VBEM-algorithm}}

\subsection{VBEM: Motivations and general guidelines}

Whether it is to estimate the posterior distribution of the parameters given the observed data and the hyperparameters, to perform hyperparameters estimation or to compare two given models, full Bayesian methods can be very useful. Here we focus on one of them called Variational Bayesian EM (VBEM) \cite{Beal2003_PhD}. It allows both to find an approximation of the posterior distribution of parameters and hidden variables (we regroup them into a single variable $\boldsymbol{W}=\left(\boldsymbol{Z},\boldsymbol{\theta},\boldsymbol{\lambda}\right)$): 
\begin{equation}
\mathcal{Q}\left(\boldsymbol{W}\right)\approx P\left(\boldsymbol{W}\mid\boldsymbol{X}_{\mathcal{O}}\right)
\end{equation}
and to compute the Evidence Lower BOund (ELBO) $\mathcal{E}$, a lower bound for the log-evidence of the data, which has generally no closed-form solution:
\begin{equation}
\mathcal{E}\left(\mathcal{Q};\boldsymbol{X}_{\mathcal{O}}\right)\leq\ln P\left(\boldsymbol{X}_{\mathcal{O}}\right)
\end{equation}
with
\begin{equation}
\mathcal{E}\left(\mathcal{Q};\boldsymbol{X}_{\mathcal{O}}\right)=\sum_{\boldsymbol{W}}\mathcal{Q}\left(\boldsymbol{W}\right)\ln\frac{P\left(\boldsymbol{W},\boldsymbol{X}_{\mathcal{O}}\right)}{\mathcal{Q}\left(\boldsymbol{W}\right)}\label{eq:VBEM_ELBO_def}
\end{equation}
and
\begin{equation}
P\left(\boldsymbol{X}_{\mathcal{O}}\right)=\sum_{W}P\left(\boldsymbol{W},\boldsymbol{X}_{\mathcal{O}}\right).\label{eq:VBEM-evidence}
\end{equation}

The goal of VBEM is to maximize $\mathcal{E}\left(\mathcal{Q};\boldsymbol{X}_{\mathcal{O}}\right)$ with respect to $\mathcal{Q}$. To do so, $\mathcal{Q}\left(\boldsymbol{W}\right)$ is usually factorized as 
\begin{equation}
\mathcal{Q}\left(\boldsymbol{W}\right)=\prod_{n=1}^{N}q_{n}\left(\boldsymbol{W}_{n}\right)
\end{equation}
where $W_{1},\dots W_{N}$ is some partition of all latent variables $\boldsymbol{W}$. It is shown that the following update rules for the $q_{n}$ distribution make the ELBO non decreasing (the notation $\left\langle f\left(x_1,\dots\right)\right\rangle _{q_{1}\left(x_1\right),\dots}$ is used for the expected value of $f\left(x_1,\dots\right)$ taking $q_{1},\dots$ as the probability distributions for $x_1,\dots$):
\begin{align}
\ln q_{n}\left(\boldsymbol{W}_{n}\right)= & \left\langle \ln P\left(\boldsymbol{W}_{1},\dots\boldsymbol{W}_{N},\boldsymbol{X}_{\mathcal{O}}\right)\right\rangle _{\left\{ q_{n^{\prime}}\left(\boldsymbol{W}_{n^{\prime}}\right)\right\} _{n^{\prime}\neq n}}\nonumber \\
 & +cst.\label{eq:VBEM_q_update_rule}
\end{align}
For the following, we define the normalized ELBO as
\begin{equation}
g^{M}\left(\mathcal{Q};\boldsymbol{X}_{\mathcal{O}}\right)=\frac{1}{M}\mathcal{E}\left(\mathcal{Q};\boldsymbol{X}_{\mathcal{O}}\right),\label{eq:VBEM-normalized-ELBO}
\end{equation}
which turns out to be well defined when $M$ tends towards infinity. This corresponds to the objective function to be maximized.

\subsection{Derivation of VBEM algorithm}

Because this will lead to VBEM algorithm that is ``easy'' to derive, we decide to take a fully factorized distribution for $\mathcal{Q}$:
\begin{align}
\mathcal{Q}\left(\boldsymbol{W}\right)= & \prod_{ij}q_{\boldsymbol{Z}_{ij}}\left(\left\{ Z_{sikj}^{m}\right\} _{msk}\right)\prod_{mkj}q_{\lambda_{kj}^{m}}\left(\lambda_{kj}^{m}\right)\nonumber \\
 & \prod_{mk}q_{\boldsymbol{\theta}_{k}^{m}}\left(\left\{ \theta_{si\mid k}^{m}\right\} _{si}\right).\label{eq:VBEM_q_tot_def}
\end{align}
Due to the symmetry with respect to $m$ of the generative process described in section \ref{subsec:The-generative-model}, VBEM will give similar definitions and update rules for $q_{\lambda_{kj}^{m}}$ and $q_{\boldsymbol{\theta}_{k}^{m}}$ for all $m$. This means that on condition that they are all initialized the same way \textendash{} which we will suppose \textendash , they will all be equals over the iterations and we can therefore ignore superscripts $m$. Note also that, for the same practical reason as for the EM algorithm (see subsection \ref{subsec:Derivation-of-EM}), hidden sources that are linked to missing data $\left\{ Z_{sikj}^{m}\right\} _{ij\notin\mathcal{O},s,k,m}$ are not excluded from $\boldsymbol{W}$. At each iteration, we update factor distributions according to equation \eqref{eq:VBEM_q_update_rule} in the following order: first, updates of parameter distributions $\left\{ q_{\lambda_{kj}^{m}}\right\} _{mkj}$ and $\left\{ q_{\boldsymbol{\theta}_{k}^{m}}\right\} _{mkj}$\footnote{It turns out that due to normalization constraint on atoms \eqref{eq:atoms normalization constraint}, those updates can be performed independently from each other, and therefore the order does not matter.}, and then updates of source distributions $\left\{ q_{\boldsymbol{Z}_{ij}}\right\} _{ij}$. The detailed calculations are provided in the supplementary material and we report here the main results. By following these guidelines, we end up with the following posterior distributions:
\begin{align}
q_{\boldsymbol{Z}_{ij}} & =\begin{cases}
\prod_{msk}\Pois\left(\bar{\ell}_{sikj}^{m}\right), & \text{ if }ij\notin\mathcal{O}\\
\DiffNom\left(MX_{ij},\left\{ \bar{\ell}_{sikj}^{m}\right\} _{msk}\right), & \text{ if }ij\in\mathcal{O}
\end{cases}\label{eq:VBEM_qa_def}\\
q_{\lambda_{kj}^{m}} & =q_{\lambda_{kj}}=\Gam\left(\hat{\alpha}_{kj},\hat{\beta}_{kj}\right),\label{eq:VBEM_qb_def}\\
q_{\boldsymbol{\theta}_{k}^{m}} & =q_{\boldsymbol{\theta}_{k}}=\Dir\left(\left\{ \hat{\alpha}_{sik}\right\} _{si}\right)\label{eq:VBEM_qc_def}
\end{align}
where $\bar{\ell}_{sikj}^{m}$ can be computed from $\hat{\alpha}_{kj},\hat{\beta}_{kj}$ and $\hat{\alpha}_{sik}$, and vice versa, leading to a EM-like alternative algorithm. A nice feature is that due to calculation simplifications, it is not necessary to explicitly compute the $\bar{\ell}_{sikj}^{m}$ variables. The resulting algorithm is described in algorithm \ref{alg:VBEM-algorithm} and is actually very closed to the EM algorithm.

Now we have a definition for the $\mathcal{Q}$ distribution, the normalized ELBO $g^{M}\left(\mathcal{Q};\boldsymbol{X}_{\mathcal{O}}\right)$ \eqref{eq:VBEM-normalized-ELBO} can be computed explicitly thanks to equation \eqref{eq:VBEM_ELBO_def}. The developped formula is given in Appendix \ref{subsec:Computation-of-ELBO}. Just know that as for the EM algorithm's objective function, it is composed of three terms that can be interpreted as a data fitting term and two regularization terms for atoms and activations.

\begin{figure*}[htb]
\centering
\begin{minipage}{.8\linewidth}
\begingroup
\removelatexerror% Nullify \@latex@error

\begin{algorithm}[H]
\BlankLine 

\KwIn{$\boldsymbol{X}$, $\mathcal{O},$ $M$}

\BlankLine 

\KwOut{$\left\{ \hat{\alpha}_{kj}\right\} $, $\left\{ \hat{\beta}_{kj}\right\} $
and $\left\{ \hat{\alpha}_{sik}\right\} $}

\BlankLine 

$\hat{\beta}_{kj}\leftarrow\beta_{\omega\left(k,j\right)}+1$

initialize $\left\{ \hat{\alpha}_{kj}\right\} $ and $\left\{ \hat{\alpha}_{sik}\right\} $

\begin{minipage}[t]{6cm}%
$\ell_{kj}\leftarrow\exp\psi\left(\hat{\alpha}_{kj}\right)/\hat{\beta}_{kj},$%
\end{minipage}%
\begin{minipage}[t]{7cm}%
$h_{sik}\leftarrow\exp\psi\left(\hat{\alpha}_{sik}\right)/\exp\psi\left(\sum_{s^{\prime}i^{\prime}}\hat{\alpha}_{s^{\prime}i^{\prime}k}\right)$%
\end{minipage}

\Repeat{convergence}{

\tcc{Compute the model and the multiplicative updates}

$\bar{\ell}_{sij}\leftarrow\sum_{k}h_{sik}\ell_{kj}$

optional: compute objective function $g^{M}\left(\mathcal{Q},\boldsymbol{X}_{\mathcal{O}}\right)$
(see Appendix \ref{subsec:Computation-of-ELBO})

$\sigma_{ij}\leftarrow\bar{\ell}_{s=0,ij}\bar{\ell}_{s=1,ij}$

$U_{sij}\leftarrow\begin{cases}
1 & \text{ if }ij\notin\mathcal{O}\\
\frac{\max\left(\left(-1\right)^{s}X_{ij},0\right)}{\bar{\ell}_{sij}}+\frac{\bar{\ell}_{1-s,ij}}{|X_{ij}|+1+\sqrt{\sigma_{ij}}\R_{|X_{ij}|+1}\left(2\sqrt{\sigma_{ij}}\right)} & \text{ if }ij\in\mathcal{O}\text{ and }M=1\\
\frac{\max\left(\left(-1\right)^{s}X_{ij},0\right)}{\bar{\ell}_{sij}}+\frac{2\bar{\ell}_{1-s,ij}}{|X_{ij}|+\sqrt{X_{ij}^{2}+4\sigma_{ij}}} & \text{ if }ij\in\mathcal{O}\text{ and }M=\infty
\end{cases}$

\begin{minipage}[t]{6cm}%
$U_{kj}^{\text{act}}\leftarrow\sum_{si}U_{sij}h_{sik}$%
\end{minipage}%
\begin{minipage}[t]{6cm}%
$U_{sik}^{\text{atoms}}\leftarrow\sum_{j}U_{sij}\ell_{kj}$%
\end{minipage}\BlankLine 

\tcc{Update parameters}

\begin{minipage}[t]{6cm}%
$\hat{\alpha}_{kj}\leftarrow\ell_{kj}U_{kj}^{\text{act}}+\alpha_{\upsilon\left(k,j\right)}$%
\end{minipage}%
\begin{minipage}[t]{6cm}%
$\hat{\alpha}_{sik}\leftarrow h_{sik}U_{sik}^{\text{atoms}}+\alpha_{\varphi\left(s,i,k\right)}$%
\end{minipage}

\begin{minipage}[t]{6cm}%
$\ell_{kj}\leftarrow\exp\psi\left(\hat{\alpha}_{kj}\right)/\hat{\beta}_{kj},$%
\end{minipage}%
\begin{minipage}[t]{7cm}%
$h_{sik}\leftarrow\exp\psi\left(\hat{\alpha}_{sik}\right)/\exp\psi\left(\sum_{s^{\prime}i^{\prime}}\hat{\alpha}_{s^{\prime}i^{\prime}k}\right)$%
\end{minipage}

\BlankLine 

}

\caption{VBEM algorithm for Skellam-SNMF. $\psi=\frac{\Gamma}{\Gamma^{\prime}}$ is the digamma function.\label{alg:VBEM-algorithm}}
\end{algorithm}

\endgroup
\end{minipage}
\end{figure*}

\subsection{Estimation of hyperparameters\label{subsec:Estimation-of-hyperparameters}}

Though it is not justified in theory, the ELBO is often used in the literature as a replacement for the log-evidence of data $\ln P\left(X_{\mathcal{O}}\right)$ \eqref{eq:VBEM-evidence} in order to perform model selection\footnote{Note that some theoretical work about the consistency of ELBO based model selection has been put forward lately \cite{Cherief-Abdellatif2019_PMLR}.} or hyperparameters estimation \cite{Cemgil2009_CIN,McGrogy2007_CSDA}. For instance, in the specific case of Poisson-NMF and given a single observed matrix $\boldsymbol{X}$, it is explained in \cite{Cemgil2009_CIN} how to infer the model order $K$ and how to alternatively run the VBEM algorithm with fixed hyperparameters, and update the hyperparameters with fixed distribution $\mathcal{Q}$, as a way to improve the estimation of the model parameters. 

Here, we focus on an alternative scenario, namely online learning of the hyperparameters given a collection of data $\left(\boldsymbol{X}^{1},\dots\boldsymbol{X}^{t},\dots\right)$, assumed to be independent and identically distributed (i.i.d.). The idea is to update the value of the hyperparameters after each run of the VBEM algorithm on a new observation, yielding an improved VBEM algorithm that is increasingly adapted to observations. To solve this problem, we first consider the case of batch estimation where the collection $\left(\boldsymbol{X}^{1},\dots\boldsymbol{X}^{T}\right)$ is fully supplied, and then explain how to switch from batch estimation to online estimation.

Assume that for each data $\boldsymbol{X}^{t}$, VBEM has provided a posterior approximation $\mathcal{Q}^{t}$, characterized by the ``posterior hyperparameters'' $\hat{\alpha}_{kj}^{t}$, $\hat{\beta}_{kj}^{t}$ and $\hat{\alpha}_{sik}^{t}$. We then wish to estimate the hyperparameters via maximization of the total normalized ELBO with respect to $\boldsymbol{\alpha}$ and $\boldsymbol{\beta}$ (in this section, $g^{M}\left(\mathcal{Q},\boldsymbol{X}\right)$ is renamed as $g^{M}\left(\mathcal{Q},\boldsymbol{X};\boldsymbol{\alpha},\boldsymbol{\beta}\right)$ ; note also that the value of $M$ plays no role in the estimation of the hyperparameters):
\begin{equation}
\hat{\boldsymbol{\alpha}},\hat{\boldsymbol{\beta}}=\argmax_{\begin{array}{c}
\boldsymbol{\alpha}>0\\
\boldsymbol{\beta}>0
\end{array}}\sum_{t}g^{M}\left(\mathcal{Q}^{t},\boldsymbol{X}^{t};\boldsymbol{\alpha},\boldsymbol{\beta}\right)\label{eq:ELBO_MAX_total_ELBO}
\end{equation}
where $g^{M}\left(\mathcal{Q}^{t},\boldsymbol{X}^{t};\boldsymbol{\alpha},\boldsymbol{\beta}\right)$ is given in appendix \ref{subsec:Computation-of-ELBO}. In order to perform this optimization, it is important to calculate the partial derivatives with respect to each hyperparameter with fixed $\mathcal{Q}^{t}$, even though $\mathcal{Q}^{t}$ is itself expressed as a function of $\boldsymbol{\alpha}$ and $\boldsymbol{\beta}$. There is no closed-form solution for this optimization, and therefore optimization algorithms must be employed. The proofs of the two following propositions can be found in the supplementary material.
\begin{prop}[Hyperparameters estimation for gamma priors]
\label{prop:ELBO-MAX-update-gamma-hyperparmeters}For $a\in\upsilon\left(\left\{ \left(k,j\right)\right\} \right)$ (i.e. for shape hyperparameters linked to activations $\boldsymbol{\lambda}$), the following update rules make the normalized ELBO non-decreasing at each iteration ($\kappa$ refers to the iteration number):
\begin{align}
\psi\left(\alpha_{a}^{\left(\kappa+1\right)}\right) & =\frac{\sum_{\left(k,j\right)\in\upsilon^{-1}\left(a\right)}\ln\beta_{\omega\left(k,j\right)}^{\left(\kappa\right)}+\gamma_{a}^{T}}{\left|\upsilon^{-1}\left(a\right)\right|},\label{eq:ELBO_MAX_gamma_alpha_d_update}\\
\beta_{b}^{\left(\kappa+1\right)} & =\frac{\sum_{\left(k,j\right)\in\omega^{-1}\left(b\right)}\alpha_{\upsilon\left(k,j\right)}^{\left(\kappa+1\right)}}{\delta_{b}^{T}},\label{eq:ELBO_MAX_gamma_beta_e_update}
\end{align}
where $\left|\upsilon^{-1}\left(a\right)\right|$ is the cardinal of inverse image $\varphi^{-1}\left(a\right)$ and where
\begin{align}
\ \gamma_{a}^{T} & =\frac{1}{T}\sum_{t=1}^{T}\sum_{\left(k,j\right)\in\upsilon^{-1}\left(a\right)}\left(\psi\left(\hat{\alpha}_{kj}^{t}\right)-\ln\hat{\beta}_{kj}^{t}\right),\label{eq:ELBO_MAX_L_dt_def}\\
\delta_{b}^{T} & =\frac{1}{T}\sum_{t=1}^{T}\sum_{\left(k,j\right)\in\omega^{-1}\left(b\right)}\frac{\hat{\alpha}_{kj}^{t}}{\hat{\beta}_{kj}^{t}}.\label{eq:ELBO_MAX_M_et_def}
\end{align}
$\psi$ is the digamma function. Its inverse can be computed using Newton's method (see Appendix C of \cite{Minka2000_TechReport}).
\end{prop}
\begin{prop}[Hyperparameters estimation for Dirichlet priors]
\label{prop:ELBO-MAX-update-dirichlet-hyperparmeters}For $a\in\varphi\left(\left\{ \left(s,i,k\right)\right\} \right)$ (i.e. for shape hyperparameters linked to atoms $\boldsymbol{\theta}$), the following update rule makes the normalized ELBO non-decreasing at each iteration ($\kappa$ refers to the iteration number):
\begin{equation}
\psi\left(\alpha_{a}^{\left(\kappa+1\right)}\right)=\frac{\sum_{sik\in\varphi^{-1}\left(a\right)}\psi\left(\sum_{s^{\prime}i^{\prime}}\alpha_{\varphi\left(s^{\prime}i^{\prime},k\right)}^{\left(\kappa\right)}\right)+\xi_{a}^{T}}{\left|\varphi^{-1}\left(a\right)\right|}\label{eq:ELBO_MAX_dirichlet_alpha_d_update1}
\end{equation}
with 
\begin{equation}
\xi_{a}^{T}=\frac{1}{T}\sum_{t=1}^{T}\sum_{sik\in\varphi^{-1}\left(a\right)}\psi\left(\hat{\alpha}_{sik}^{t}\right)-\psi\left(\sum_{s^{\prime}i^{\prime}}\hat{\alpha}_{s^{\prime}i^{\prime}k}^{t}\right).\label{eq:ELBO_MAX_N_dt_def}
\end{equation}
\end{prop}
It is quite simple to switch to online estimation since quantities $\gamma_{a}^{T}$, $\delta_{b}^{T}$ and $\xi_{a}^{T}$ can be computed recursively:
\begin{align}
\gamma_{a}^{T+1} & =\left(1-c\right)\gamma_{a}^{T}+c\sum_{kj\in\upsilon^{-1}\left(d\right)}\left(\psi\left(\hat{\alpha}_{kj}^{T}\right)-\ln\hat{\beta}_{kj}^{T}\right),\label{eq:ELBO_MAX_Ld_t_recursif}\\
\delta_{b}^{T+1} & =\left(1-c\right)\delta_{b}^{T}+c\sum_{kj\in\omega^{-1}\left(e\right)}\frac{\hat{\alpha}_{kj}^{T}}{\hat{\beta}_{kj}^{T}},\label{eq:ELBO_MAX_M_et_recursif}\\
\xi_{a}^{T+1} & =\left(1-c\right)\xi_{a}^{T}+c\underset{{\scriptstyle sik\in\varphi^{-1}\left(a\right)}}{\sum}\psi\left(\hat{\alpha}_{sik}^{T}\right)-\psi\left(\sum_{s^{\prime}i^{\prime}}\hat{\alpha}_{s^{\prime}i^{\prime}k}^{T}\right)\label{eq:ELBO_MAX_N_dt_recursif}
\end{align}
with $c=\frac{1}{T+1}$. Therefore, on the condition that a record of those three quantities is kept, as well as the number $T$, each time a new observation is provided, one can run VBEM, update $\boldsymbol{\gamma}$, $\boldsymbol{\delta}$ and $\boldsymbol{\xi}$ and finally update $\boldsymbol{\alpha}$ and $\boldsymbol{\beta}$ using propositions \ref{prop:ELBO-MAX-update-gamma-hyperparmeters} and \ref{prop:ELBO-MAX-update-dirichlet-hyperparmeters}.

Note that it may be interesting to fix the value $c$ once for all, independent of $T$, since it has two advantages. First, it prevents overfitting in the early stages of the process ($T$ small), when there is still little data to learn from, and moreover when estimation of $\hat{\alpha}_{kj}^{t\leq T}$ and $\hat{\alpha}_{sik}^{t\leq T}$ might be poor due to inappropriate values for the hyperparameters. Second, it gradually erases the contributions of past observations, allowing the process to be resilient in case observations $\left(\boldsymbol{X}^{1},\dots\boldsymbol{X}^{t},\dots\right)$ were not strictly identically distributed. $c$ can then be seen as a \emph{learning rate}, and be set to a small value (\emph{e.g. $c=0.02$).} Finally, $\gamma_{a}$, $\delta_{b}$ and $\xi_{a}$ can be initialized using equations \eqref{eq:ELBO_MAX_L_dt_def}, \eqref{eq:ELBO_MAX_M_et_def} and \eqref{eq:ELBO_MAX_N_dt_def}, with $T=1$, $\hat{\alpha}_{kj}=\alpha_{\nu\left(k,j\right)}$, $\hat{\beta}_{kj}=\beta_{\omega\left(k,j\right)}$ and $\hat{\alpha}_{sik}=\alpha_{\varphi\left(s,i,k\right)}$.

\section{Experimental studies\label{sec:Experimental-studies}}

We conduct several experiments in order to both study the intrinsic performance and characteristics of our estimation algorithms on synthetic data and to evaluate Skellam-SNMF for automatic clustering on real data in comparison with the original SNMF algorithm. Acronyms and other information about the algorithms that will be used in this section are presented in Table \ref{tab:SNMF-algorithms}. All our Skellam-SNMF algorithms are implemented using the Wonterfact python package \cite{Fuentes2020_web}. This package, developed by the main author of this paper, allows the design of any kind of tensor factorization model, included Skellam-SNMF, with an automatic derivation of the EM or the VBEM algorithm. The code to reproduce all the experiments in this section can be found in the "jupyter" directory of the Wonterfact repository.

\begin{table}[tbh]
\caption{SNMF algorithms.\label{tab:SNMF-algorithms} Note that $K$-means can be interpreted as SNMF with only binary entries for the activations. }

\begin{centering}
\begin{tabular}{l>{\raggedright}p{0.27\columnwidth}>{\raggedright}p{0.4\columnwidth}}
\toprule 
\addlinespace[1pt]
\multicolumn{1}{c}{Acronym} & \multicolumn{1}{c}{Description} & \multicolumn{1}{c}{Remark}\tabularnewline\addlinespace[1pt]
\midrule
\addlinespace[1pt]
Sk$_{M}$ & Skellam-SNMF with EM algorithm  & $M=1$ or $\infty$ (see section \ref{subsec:The-generative-model})\tabularnewline
Sk$_{M}$-VB & Skellam-SNMF with VBEM algorithm & $M=1$ or $\infty$ (see section \ref{subsec:The-generative-model})\tabularnewline
Ding'10 & Original SNMF algorithm \cite{Ding2010} & We used implementation \cite{Thurau2014_github} \tabularnewline
$K$-means & Classic $K$-means algorithm & We used implementation \cite{Thurau2014_github} \tabularnewline
\bottomrule
\end{tabular}
\par\end{centering}
\end{table}

\subsection{Parameter estimation on synthetic data\label{subsec:Parameter-estimation-on}}

In the first experiment, we generate integer data according to the generative process presented in subsection \ref{subsec:The-generative-model} with $M=1$ and we study the performance of Sk$_{1}$ and Sk$_{1}$-VB (see Table \ref{tab:SNMF-algorithms}) in the ideal case where the hyperparameters used to generate data are known. In order to obtain easily interpretable and visualizable results, we decide to generate a large number ($J=5000)$ of $3$-dimensional data ($I=3$) with two components ($K=2$). Mapping functions $\varphi$, $\nu$ and $\omega$ are chosen such that each coefficient of atoms $\boldsymbol{\theta}$ has its own hyperparameter and that activation's hyperparameters only depend on component $k$:
\begin{align}
\left\{ \theta_{si\mid k}\right\} _{si} & \sim\Dir\left(\left\{ \alpha_{sik}\right\} _{si}\right),\\
\lambda_{kj} & \sim\Gam\left(\alpha_{k},\beta_{k}\right).
\end{align}
In order to set the values of the hyperparameters, we randomly draw shapes $\alpha_{sik}$ and manually set shapes $\alpha_{k}$, according to two target levels of prior uncertainty for the parameters. Rates $\beta_{k}$ are set such that mean value of activations is $300$, leading to observations in the order of magnitude of a hundred. For each of the uncertainty level, we add a ``low variance'' option: activating this option corresponds to setting for each dimension $i$ and component $k$ $\alpha_{sik}=\epsilon$ for $s=0$ or $1$, where $\epsilon$ is some small value. Doing so insure that either $\theta_{s=0,i\mid k}$ or $\theta_{s=1,i\mid k}$ is closed to zero and thus that variance of the Skellam hidden sources $Z_{ikj}=Z_{s=0,ikj}-Z_{s=1,ikj}\sim\Skell\left(\theta_{s=0,i\mid k}\lambda_{kj},\theta_{s=1,i\mid k}\lambda_{kj}\right)$ is minimal. A low variance of the hidden sources can be interpreted as a low level of noise in the observed data. The values of the hyperparameters are summarized in Table \ref{tab:synth_data_hyperparam}.

\begin{table}[tbh]
\caption{Values of hyperparameters according to the target level of prior uncertainty \label{tab:synth_data_hyperparam}. If low variance option is activated, then $\min_{s}\alpha_{sik}$ is set to $0.02$.}

\centering{}%
\begin{tabular}{ll}
\toprule 
\multicolumn{1}{c}{low uncertainty} & \multicolumn{1}{c}{high uncertainty}\tabularnewline
\midrule
\addlinespace[1pt]
$\alpha_{sik}\in\left[1\ 10\right]$ & $\alpha_{sik}\in\left[0.5\ 1\right]$\tabularnewline\addlinespace[1pt]
\addlinespace[1pt]
\addlinespace[1pt]
$\alpha_{1}=5,\alpha_{2}=50$ & $\alpha_{1}=0.8,\alpha_{2}=0.5$\tabularnewline\addlinespace[1pt]
\addlinespace[1pt]
\addlinespace[1pt]
$\beta_{k}=\alpha_{k}/300$ & $\beta_{k}=\alpha_{k}/300$\tabularnewline\addlinespace[1pt]
\bottomrule
\addlinespace[1pt]
\end{tabular}
\end{table}

For each set of values for the hyperparameters, we iterate $50$ times the drawing of observations according to the generative process and the running of Sk$_{1}$ and Sk$_{1}$-VB. Concerning the initialization, we set $\hat{\theta}_{si\mid k}^{\left(0\right)}=\alpha_{sik} / \sum_{si}\alpha_{sik}$ and $\hat{\theta}_{kj}^{\left(0\right)}=\alpha_{k}/(1+\beta_k)$ for Sk$_{1}$ and $\hat{\alpha}_{sik}^{\left(0\right)}=\alpha_{sik}$, and $\hat{\alpha}_{kj}^{\left(0\right)}=\alpha_{k}$ for Sk$_{1}$-VB. After convergence, we decide for Sk$_{1}$-VB to take the mean value of estimated posterior distributions $\hat{q}_{\mathbb{\theta}_{k}}$ and $\hat{q}_{\lambda_{kj}}$ as the parameter estimates. In order to assess the quality of the estimation of both atoms and activations, we suggest to compute the mean square error (mse) on estimated parameters of the Skellam hidden sources $Z_{ikj}$. Besides, so we have interpretable results, we can compute separate mse for the expectation and the variance of those hidden sources:
\begin{align}
\textrm{mse}_{m} & =\frac{\sum_{ikj}\left(m_{ikj}-\hat{m}_{ikj}\right)^{2}}{I\times K\times J},\\
\textrm{mse}_{v} & =\frac{\sum_{ikj}\left(v_{ikj-}\hat{v}_{ikj}\right)^{2}}{I\times K\times J},
\end{align}
with $m_{ikj}=W_{ik}\lambda_{kj}=\left(\theta_{s=0,i\mid k}-\theta_{s=1,i\mid k}\right)\lambda_{kj}$ and $v_{ikj}=\left(\theta_{s=0,i\mid k}+\theta_{s=1,i\mid k}\right)\lambda_{kj}$ (same definitions for $\hat{m}_{ikj}$ and $\hat{v}_{ikj}$).

Both algorithms are compared to the ``dummy'' algorithm consisting in taking the mean values of prior distribution as the estimation for the parameters. Results are shown in Table \ref{tab:results_sk_1_synth_data} from which several conclusions can be drawn. First of all, it can be noticed that results given by Sk$_{1}$ and Sk$_{1}$-VB are always of the same order of magnitude: at this point, we can claim that Sk$_{1}$-VB presents no significant advantage over Sk$_{1}$ for parameter estimation with known hyperparameters. Second, when the low variance option is activated, both Sk$_{1}$ and Sk$_{1}$-VB give good results with low values of mean and standard deviation. The difference with the dummy algorithm is particularly important in the high uncertainty scenario. This is expected since the greater the uncertainty on the \emph{a priori} value of the parameters, the more crucial the observed data are in order to give a good estimation. A more surprising result is the poor quality of the parameter estimates when the low variance option is not activated. A probable explanation is that in objective functions of both Sk$_{1}$ and Sk$_{1}$-VB, the data fitting term prevails over the prior terms, meaning that these algorithms prefer minimizing the reconstruction error, \emph{i.e.} the variance of the Skellam hidden sources, than complying to the priors.

\begin{table*}[t]
\begin{centering}
\caption{Mean $\mid$ standard deviation of the different metrics on $50$ runs with respect to the prior uncertainty level and the low variance option \label{tab:results_sk_1_synth_data}}
\par\end{centering}
\centering{}%
\begin{tabular}{llrrrr}
\toprule 
metric & \multicolumn{1}{c}{algorithm} & \multicolumn{1}{c}{low uncertainty} & \multicolumn{1}{c}{low uncertainty, low variance} & \multicolumn{1}{c}{high uncertainty} & \multicolumn{1}{c}{high uncertainty, low variance}\tabularnewline
\midrule
\multirow{3}{*}{$mse_{m}$ ($\times10^{3}$)} & Dummy & $1.01\mid0.54$ & $1.95\mid0.52$ & $15.74\mid06.76$ & $31.57\mid8.33$\tabularnewline
 & Sk$_{1}$ & $\boldsymbol{0.66}\mid0.49$ & $0.18\mid0.03$ & $\boldsymbol{5.99}\mid12.27$ & $0.73\mid0.97$\tabularnewline
 & Sk$_{1}$-VB & $0.90\mid0.68$ & $\boldsymbol{0.12}\mid0.03$ & $6.63\mid13.23$ & $\boldsymbol{0.57}\mid1.34$\tabularnewline
\midrule
\multirow{3}{*}{$mse_{v}$ ($\times10^{3}$) } & Dummy & $\boldsymbol{1.75}\mid0.36$ & $1.98\mid0.51$ & $25.42\mid05.77$ & $32.63\mid7.27$\tabularnewline
 & Sk$_{1}$ & $3.09\mid2.31$ & $0.19\mid0.05$ & $13.82\mid13.38$ & $\boldsymbol{1.37}\mid1.99$\tabularnewline
 & Sk$_{1}$-VB & $2.84\mid2.35$ & $\boldsymbol{0.13}\mid0.06$ & $\boldsymbol{13.45}\mid12.79$ & $1.43\mid3.43$\tabularnewline
\bottomrule
\end{tabular}
\end{table*}

\subsection{Online hyperparameter estimation on synthetic data}

\begin{figure}[tbh]
\subfloat[]{\begin{centering}
\includegraphics[width=0.95\columnwidth]{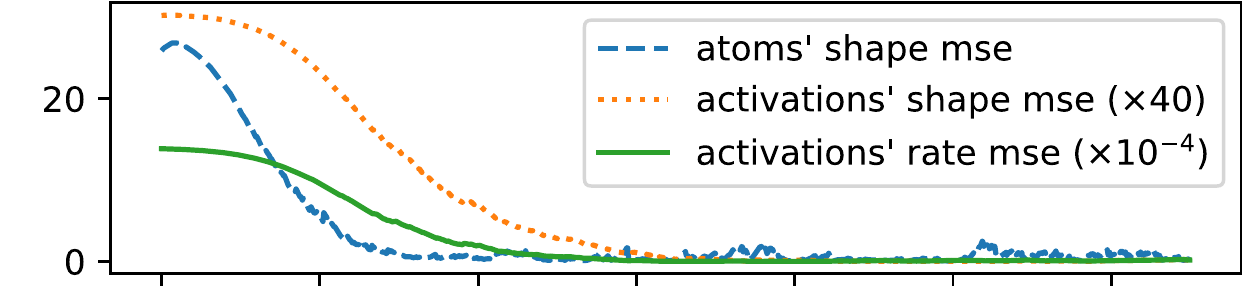}
\par\end{centering}

}

\subfloat[]{\begin{centering}
\includegraphics[width=0.95\columnwidth]{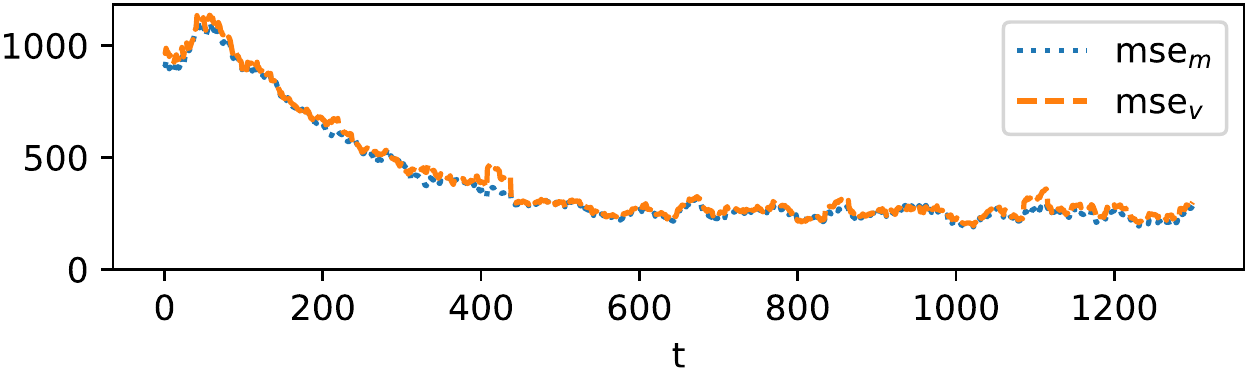}
\par\end{centering}

}\caption{\label{fig:hyperparam_estimation}Hyperparameters (a) and parameters (b) estimation error is globally decreasing with respect to the number $t$ of analyzed data. The best permutation of components $k$ is found before the computations of each mse. $\text{mse}_{m}$ and $\text{mse}_{v}$ are averaged over $30$ consecutive results in order to smooth out the performance variability.}

\end{figure}

In this experiment, we show that Sk$_{1}$-VB along with hyperparameter estimation can be used to perform unsupervised learning. As a proof of concept, we decide to generate a dataset $\left\{ \boldsymbol{X}^{1},\dots\boldsymbol{X}^{t},\dots\boldsymbol{X}^{T}\right\} $ according to the same ``low uncertainty, low variance'' scenario as in previous subsection, with always the same hyperparameters. Contrary to the previous experiment, those hyperparameters are unknown and to be estimated. To do so, we first initialize hyperparameters for the Sk$_{1}$-VB algorithm with neutral values (all shape hyperparameters are set to $1$ and rate hyperparameters are set to $0.001$), and then for each data $\boldsymbol{X}^{t}$, we run Sk$_{1}$-VB until convergence and then update hyperparameters as described in section \ref{subsec:Estimation-of-hyperparameters}. In figure \ref{fig:hyperparam_estimation}, it is showed how the mse of estimated hyperparameters with respect to the number of analyzed data is globally decreasing in a first learning stage, and then globally stable and close to $0$ at convergence. It is also showed that while the estimated hyperparmeters are getting closer to the ground truth, mse of the parameters (that is $\textrm{mse}_{m}$ and $\textrm{mse}_{v}$ as defined in the previous subsection) are also getting better and better. This proves that our parameter estimation algorithm can automatically improves itself as data are collected.

Note that we ran this experiment five times, with several random values for the ground truth hyperparameters, and those promising results were achieved only twice out of the five runs. In the other cases, the process degenerated into some vicious circle, where the more biased the hyperparameter estimates, the more biased the estimation of the posterior distributions for the parameters given a new observed data, and \emph{vice versa}, leading to quite bad results. We have not yet conducted any research in order to better understand and circumvent this issue, hence the qualification of this experience as a proof of concept. We believe however that two leads should be explored. The first one would be to supervise the VBEM algorithm in the early stages of this process. It would assure that the parameters posterior distributions are well estimated at the beginning, and then prevent the hyperparameters estimation from taking a wrong direction. The second one would be to attenuate somehow the role of the parameters prior distribution during the last iterations of each VBEM algorithm. Doing so would prevent priors to take precedence over the data in case they were too strong.

\subsection{Difference between Sk\texorpdfstring{$_{1}$}{1} and Sk\texorpdfstring{$_{\infty}$}{inf} for SNMF}

\begin{figure}[tbh]
\begin{centering}
\subfloat[Estimated vs ground truth activations for $k=1$.]{\centering{}\includegraphics[width=0.95\columnwidth]{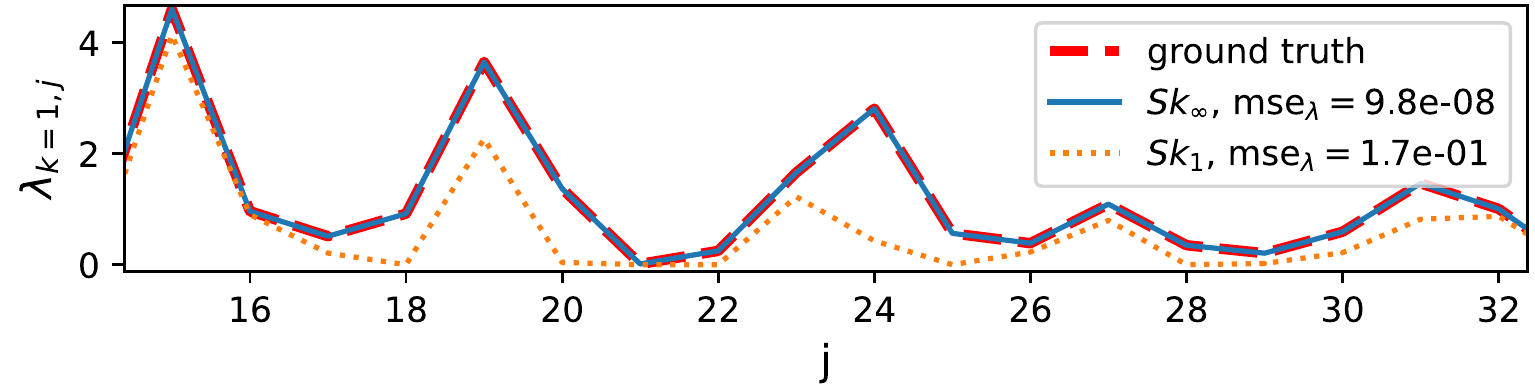}}
\par\end{centering}
\begin{centering}
\subfloat[Ratio between two consecutive values of the objective function. The fastest it gets to $1$, the fastest the convergence.]{\centering{}\includegraphics[width=0.95\columnwidth]{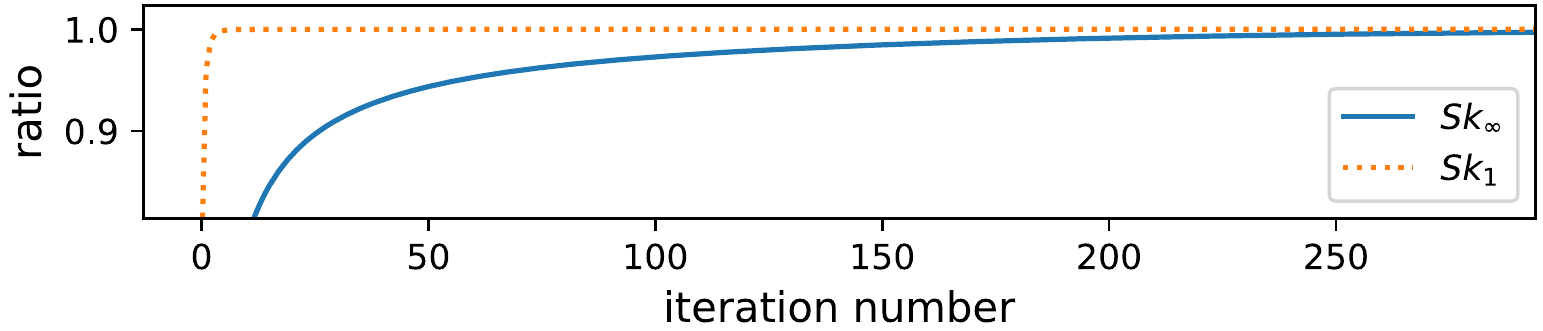}}\caption{\label{fig:Comparison-between-Sk}Comparison between Sk$_{1}$ and Sk$_{\infty}$ in a task of supervised SNMF.}
\par\end{centering}
\end{figure}

\begin{figure}[tbh]
\begin{centering}
\subfloat[]{\centering{}\includegraphics[width=0.7\columnwidth]{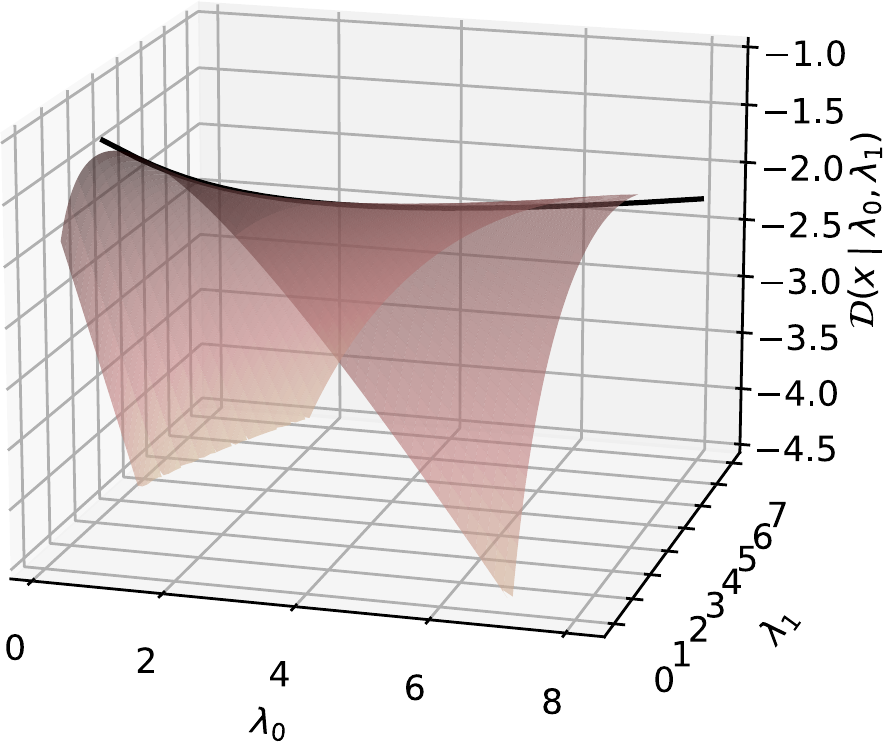}}
\par\end{centering}
\centering{}\subfloat[]{\centering{}\includegraphics[width=0.7\columnwidth]{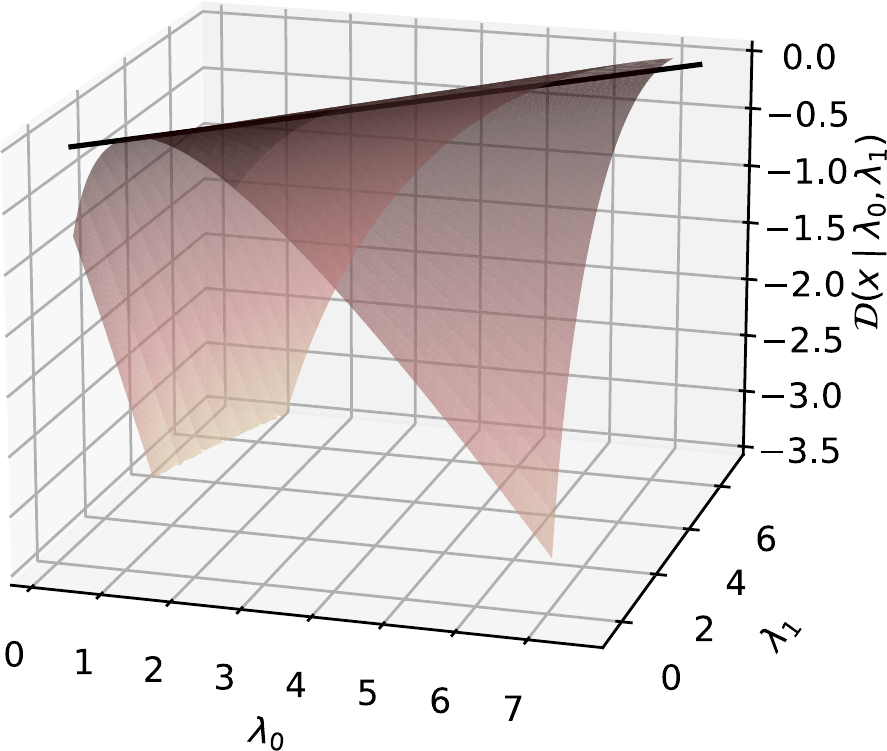}}\caption{\label{fig:3-dimensional-plots-of}3-dimensional plots of objective function basic terms of Sk$_{1}$ (a) vs Sk$_{\infty}$ (b) when observed data $x=1$ (no prior).}
\end{figure}

The goal of this subsection is to better understand the concrete differences between Sk$_{1}$ and Sk$_{\infty}$ in SNMF problems, besides the fact that one is theoretically supposed to process only integer data and the other only real-valued data. To this aim, we generate a real matrix $\boldsymbol{X}$ as the product of a ground truth atoms matrix $\boldsymbol{W}$ and a ground truth activations matrix $\boldsymbol{\lambda}$, with no addition of noise, and we ask Sk$_{1}$ and Sk$_{\infty}$ to give an estimate $\hat{\boldsymbol{\lambda}}$ of $\boldsymbol{\lambda}$, given $\boldsymbol{W}$, meaning that atoms $\boldsymbol{\theta}$ are fixed and set up to ($\propto$ is for ``proportional to'')
\begin{equation}
\theta_{si\mid k}\propto\frac{\left|W_{ik}\right|+\left(-1\right)^{s}W_{ik}}{2}.
\end{equation}
We run our algorithms with no prior on parameters. The dimensions used to generate data are $I=10$, $K=3$, $J=100$, and ground truth atoms and activations are randomly drawn. In figure \ref{fig:Comparison-between-Sk}, the estimated $\hat{\boldsymbol{\lambda}}$ compared to $\boldsymbol{\lambda}$ are plot for a given $k$, as well as the ratio between two consecutive values of the objective function with respect to the iteration number. Two simple conclusions can be made. The first one is that Sk$_{1}$ gives biased values for the activations \textendash{} especially when value of $\lambda_{kj}$ is low \textendash{} while Sk$_{\infty}$ provides an exact estimation. The second one is that the convergence of Sk$_{\infty}$ is quite slow compared to that of Sk$_{1}$. These two characteristics can be explained by visualizing on figure \ref{fig:3-dimensional-plots-of} the shape of the objective function basic terms, which are Skellam log-likelihood $\log P_{\text{skel}}\left(x\mid\lambda_{0},\lambda_{1}\right)$ (see equation \eqref{eq:Skellam-distribution}) for Sk$_{1}$ and $-\mathcal{D}\left(x\mid\lambda_{0},\lambda_{1}\right)$ (see equation \eqref{eq:EM-log-likelihood-limit}) for Sk$_{\infty}$. The bias in activations' estimate is due to the fact that $\log P_{\text{skel}}\left(x\mid\lambda_{0},\lambda_{1}\right)$ gets higher when $\lambda_{0}-\lambda_{1}$ indeed gets closed to $x$ but also when $\lambda_{0}+\lambda_{1}$ is minimal. On the opposite, $-\mathcal{D}\left(x\mid\lambda_{0},\lambda_{1}\right)$ is always maximal if $x=\lambda_{0}-\lambda_{1}$, no matter the value of $\lambda_{0}+\lambda_{1}$. As a drawback, the shape of $\mathcal{D}\left(x\mid\lambda_{0},\lambda_{1}\right)$ becomes rather flat when both $\lambda_{0}$ and $\lambda_{1}$ go away from $0$, which can explain the slowness of the convergence. Hopefully, there exists acceleration methods for the EM algorithm. We have implemented the parabolic EM algorithm \cite{Berlinet2012_CSDA}, which showed a drastic acceleration of the convergence.

Note that in order to emphasize the difference between the two algorithms, we drew low random values for $\lambda_{kj}$, and thus for matrix $\boldsymbol{X}$: for large values of observed data, Sk$_{1}$ tends to behave like Sk$_{\infty}$ even for non-integer data.

\subsection{Automatic clustering on real data}

The last experiment we conduct aims at comparing our new Skellam-SNMF technique with the classic SNMF with the Euclidean distance as the objective function. We chose to do so in a simple automatic clustering task, since it is the application that has been originally proposed \cite{Ding2010}. The idea behind using SNMF for this task is that atoms can represent the centroids of the clusters while the activations can account for the cluster membership of the observed samples. All datasets used in this experiments are from the UCI repository \cite{Dua2019_web} and are composed of real-valued data. They are summarized in Table \ref{tab:dataset}, and include the datasets \emph{Ionosphere} and \emph{Wave}, that have already been used in \cite{Ding2010}.

\begin{table}[tbh]
\caption{Dataset description \label{tab:dataset}.}

\centering{}%
\begin{tabular}{lcccc}
\toprule 
 & Ionosphere & Wave & Image & Shuttle\tabularnewline
\midrule
\# instances ($J$) & 351 & 5000 & 2310 & 14500\tabularnewline
\# attributes ($I$) & 34 & 21 & 19 & 9\tabularnewline
\# classes ($K$) & 2 & 3 & 7 & 7\tabularnewline
\bottomrule
\end{tabular}
\end{table}
We run $5$ algorithms on those data: Ding'10, Sk$_{\infty}$ and Sk$_{\infty}$-VB with all prior shapes set to $1$ and all prior rates for the activations set to $0.001$ (which allows to regularize activations and to prevent them from tending towards $\infty$), $K$-means, and finally the ``dummy'' algorithm consisting in assigning all the samples in a single class. After convergence, the value of $k$ for which $\hat{\lambda}_{kj}$ is maximum defines the class of sample $j$. As in \cite{Ding2010}, the performance measure is the clustering accuracy: first, the confusion matrix is computed, and then the columns are reordered so that the sum of the diagonal is maximal. This sum defines the metric, and represents the percentage of samples correctly clustered. Each algorithm is run $100$ times, with random initialization, and mean and standard deviation are reported in Table \ref{tab:clustering-results}.
\begin{table}[tbh]
\caption{Mean | standard deviation of the clustering accuracy ($\%$) on $100$ runs. \label{tab:clustering-results}.}

\centering{}%
\begin{tabular}{lrrrr}
\toprule 
 & \multicolumn{1}{c}{Ionosphere} & \multicolumn{1}{c}{Wave} & \multicolumn{1}{c}{Image} & \multicolumn{1}{c}{Shuttle}\tabularnewline
\midrule
dummy & $64.1\mid0.0$ & $33.9\mid00.0$ & $14.3\mid0.0$ & $\boldsymbol{79.2}\mid00.0$\tabularnewline
$K$-means & $\boldsymbol{70.8}\mid1.6$ & $50.2\mid00.0$ & $\boldsymbol{52.2}\mid5.0$ & $60.8\mid09.9$\tabularnewline
Ding'10 & $58.7\mid4.8$ & $61.9\mid10.1$ & $46.9\mid5.0$ & $30.0\mid05.1$\tabularnewline
Sk$_{\infty}$ & $70.6\mid0.6$ & $\boldsymbol{64.5}\mid10.8$ & $48.2\mid5.6$ & $53.1\mid12.4$\tabularnewline
Sk$_{\infty}$-VB & $70.7\mid0.0$ & $64.1\mid10.2$ & $50.7\mid3.0$ & $36.8\mid07.4$\tabularnewline
\bottomrule
\end{tabular}
\end{table}
From these results, the following conclusions can be made.
\begin{enumerate}[wide, labelindent=0pt]
\item Sk$_{\infty}$ and Sk$_{\infty}$-VB always outperform Ding'10. This seems to show that the divergence $\mathcal{D}\left(x\mid\lambda_{0},\lambda_{1}\right)$ is a good alternative to the Euclidean distance for real-valued data, the same way the Kullback-Leibler divergence is also a good alternative for nonnegative data in some applications \cite{Fitzgerald2009_ISSC}. This is our main experimental result.
\item We do not share the same conclusion as in Ding \emph{et. al.} \cite{Ding2010}, where they find that matrix factorization models are better than $K$-means. On the contrary, in our experiment, $K$-means outperforms SNMF algorithms on $3$ datasets over $4$. Note that on the \emph{Ionosphere} dataset and for the $K$-means algorithm, we report a mean accuracy of $70.8\%$ when Ding \emph{et. al. }report $42.2\%$. We suspect that it might be either a misprint or an error in the setting of $K$. Indeed, this result in addition to being very different from ours, does not make sense: in a $2$ classes automatic clustering scenario, accuracy is necessarily above $50\%$. However, we agree that SNMF algorithms can compete with $K$-means.
\item In the \emph{Shuttle} dataset, the number of sample per class is very unbalanced, with $79.2 \%$ of the samples belonging to a single class. In this case, neither $K$-means nor other matrix factorization models seems to be relevant as is since the dummy algorithm gives better performances. This is expected as soon as we minimize the global reconstruction error: classes with too few representatives will not affect that much the objective function.
\item Sk$_{\infty}$ and Sk$_{\infty}$-VB gives similar results, as in our very first experiment (subsection \ref{subsec:Parameter-estimation-on}). This confirms the fact that VBEM algorithm does not seem to outperform EM algorithm for the parameters estimation task. Though, we observe that the standard deviation of the results is always slightly less in Sk$_{\infty}$-VB than in Sk$_{\infty}$ (with even a $0$ standard deviation for the \emph{Ionosphere} dataset). An explanation may be found in the role of all shape hyperparameters (set to $1$ in this experiment): in the EM algorithm they play no role whatsoever in the computation of the objective function \eqref{eq:EM-full-objective-function}, whereas they do in VBEM's objective function \eqref{eq:VBEM_ELBO_tot}. Sk$_{\infty}$-VB has then an extra regularization term compared to Sk$_{\infty}$ , which can reduce the variability of the results with respect to the initialization. 
\end{enumerate}

\section{Conclusions\label{sec:Conclusions}}

\subsection{Main contributions}

We have put forward a probabilistic model called Skellam-SNMF in order to address the SNMF problem. This model is an extension of the Poisson-NMF where the NMF with the KL divergence is interpreted as a statistical inference problem using Poisson-distributed latent sources. Skellam-SNMF is based on the Skellam distribution, and allowed us to introduce a new divergence between a real number $x$ and two nonnegative parameters. This divergence can be interpreted as a generalization of KL divergence for real valued data. Its introduction is in our opinion the main contribution of this paper since it can be used as an alternative to the standard Euclidean distance in many other domains.

We have derived two algorithms in order to estimate the parameters of Skellam-SNMF, the EM and the VBEM algorithms, and we have also seen how to estimate the hyperparameters. It has been showed in the experiments that VBEM did not seems to give better estimates than EM given fixed hyperparameters, but could improve itself as it analyzed new data due to online estimation of the parameters latent prior distribution. This feature can be interpreted as a way to conjugate blind data processing methods (as matrix factorizations are often considered) and automatic learning. We consider it as our second main contribution.

Finally, we have shown that Skellam-SNMF could compete with the original SNMF model with Euclidean distance in a simple task of automatic clustering. This gives then a alternative algorithm to be tested for all applications that needs SNMF.

\subsection{Forthcoming work}

Two features have been put forward in Skellam-SNMF without being tested due to the lack of space. The first one is the ability to deal with missing data, which could be used for data restoration applications such as inpainting or for prediction problems like movie or music recommendation. The second one is the automatic estimation of model order $K$ which is possible by comparing the ELBO values of two competing models. This will be done in future work.

Furthermore, a generalization of Skellam-SNMF for any semi-nonnegative tensor factorization model, the same way Generalized Coupled Tensor Factorization \cite{Yilmaz2011_NIPS} is a generalization of Poisson-NMF, is currently under publication. This generalization has already been developed and implemented in the Wonterfact package \cite{Fuentes2020_web}. A technical report containing the underlying theory can be found in the repository of this package.

Finally, as we have already mentioned, the ability of self-improvement via the online estimation of parameters prior distribution seems very promising. We wish to further study this feature, find strategies in order to not let the process degenerate, and test it in real applications.

\appendices{}

\section{Expression of ELBO}\label{subsec:Computation-of-ELBO}
Supposing that $\mathcal{Q}$ is defined as in equations \eqref{eq:VBEM_q_tot_def}
to \eqref{eq:VBEM_qc_def}, the normalized ELBO can be computed as:

\begin{equation}
g^{M}\left(\mathcal{Q},\boldsymbol{X}_{\mathcal{O}}\right)=g_{\boldsymbol{Z}}^{M}\left(\mathcal{Q},\boldsymbol{X}_{\mathcal{O}}\right)+\sum_{kj}g_{\lambda_{kj}}\left(\mathcal{Q}\right)+\sum_{k}g_{\boldsymbol{\theta}_{k}}\left(\mathcal{Q}\right)\label{eq:VBEM_ELBO_tot}
\end{equation}
with (the definitions of $\bar{\ell}_{sij}$ and $\sigma_{ij}$ can
be found in Algorithm \ref{alg:VBEM-algorithm})
\begin{equation}
\begin{aligned} & g_{\boldsymbol{Z}}^{1}\left(\mathcal{Q}\right)=-\sum_{kj}\frac{\hat{\alpha}_{kj}}{\hat{\beta}_{kj}}+\sum_{ij\in\mathcal{O}}\left(\ln\frac{_{0}F_{1}\left(\left|X_{ij}\right|+1,\sigma_{ij}\right)}{\Gamma\left(\left|X_{ij}\right|+1\right)}\right.\\
 & \left.+\sum_{s}\max\left(\left(-1\right)^{s}X_{ij},0\right)\ln\bar{\ell}_{sij}\right)+\sum_{ij\notin\mathcal{O}}\sum_{s}\bar{\ell}_{sij},
\end{aligned}
\label{eq:VBEM-ELBO-Z}
\end{equation}
\begin{equation}
\begin{aligned} & g_{\boldsymbol{Z}}^{\infty}\left(\mathcal{Q}\right)=-\sum_{kj}\frac{\hat{\alpha}_{kj}}{\hat{\beta}_{kj}}+\sum_{ij\in\mathcal{O}}\left(\right.\\
 & +\left|X_{ij}\right|\ln\left(\frac{\left|X_{ij}\right|+\sqrt{X_{ij}^{2}+4\sigma_{ij}}}{2}\right)-\sqrt{X_{ij}^{2}+4\sigma_{ij}}\\
 & \left.+\sum_{s}\max\left(\left(-1\right)^{s}X_{ij},0\right)\ln\bar{\ell}_{sij}\right)+\sum_{ij\notin\mathcal{O}}\sum_{s}\bar{\ell}_{sij},
\end{aligned}
\label{eq:VBEM-ELBO-Z-inf}
\end{equation}
\begin{equation}
\begin{aligned} & g_{\lambda_{kj}}\left(\mathcal{Q}\right)=\alpha_{\upsilon\left(k,j\right)}\ln\frac{\beta_{\omega\left(k,j\right)}}{\hat{\beta}_{kj}}+\hat{\alpha}_{kj}\left(1-\frac{\beta_{\omega\left(k,j\right)}}{\hat{\beta}_{kj}}\right)\\
 & -\psi\left(\hat{\alpha}_{kj}\right)\left(-\alpha_{\upsilon\left(k,j\right)}\right)-\ln\frac{\Gamma\left(\alpha_{\upsilon\left(k,j\right)}\right)}{\Gamma\left(\hat{\alpha}_{kj}\right)},
\end{aligned}
\label{eq:VBEM-ELBO-activations}
\end{equation}
\begin{equation}
\begin{aligned} & g_{\boldsymbol{\theta}_{k}}\left(\mathcal{Q}\right)=\ln\frac{\Gamma\left(\sum_{si}\alpha_{\varphi\left(s,i,k\right)}\right)}{\Gamma\left(\sum_{si}\hat{\alpha}_{sik}\right)}-\sum_{si}\ln\frac{\Gamma\left(\alpha_{\varphi\left(s,i,k\right)}\right)}{\Gamma\left(\hat{\alpha}_{sik}\right)}\\
 & -\sum_{si}\left(\hat{\alpha}_{sik}-\alpha_{\varphi\left(s,i,k\right)}\right)\left(\psi\left(\hat{\alpha}_{sik}\right)-\psi\left(\sum_{s^{\prime}i^{\prime}}\hat{\alpha}_{s^{\prime}i^{\prime}k}\right)\right).
\end{aligned}
\label{eq:VBEM-ELBO-atoms}
\end{equation}

Detailed calculations can be found in the supplementary material.

\bibliographystyle{IEEEtran}
\bibliography{references}

\begin{thebibliography}{10}
\providecommand{\url}[1]{#1}
\csname url@samestyle\endcsname
\providecommand{\newblock}{\relax}
\providecommand{\bibinfo}[2]{#2}
\providecommand{\BIBentrySTDinterwordspacing}{\spaceskip=0pt\relax}
\providecommand{\BIBentryALTinterwordstretchfactor}{4}
\providecommand{\BIBentryALTinterwordspacing}{\spaceskip=\fontdimen2\font plus
\BIBentryALTinterwordstretchfactor\fontdimen3\font minus
  \fontdimen4\font\relax}
\providecommand{\BIBforeignlanguage}[2]{{%
\expandafter\ifx\csname l@#1\endcsname\relax
\typeout{** WARNING: IEEEtran.bst: No hyphenation pattern has been}%
\typeout{** loaded for the language `#1'. Using the pattern for}%
\typeout{** the default language instead.}%
\else
\language=\csname l@#1\endcsname
\fi
#2}}
\providecommand{\BIBdecl}{\relax}
\BIBdecl

\bibitem{Comon2010_book}
P.~Comon and C.~Jutten, \emph{{Handbook of Blind Source Separation: Independent
  Component Analysis and Applications}}, 1st~ed.\hskip 1em plus 0.5em minus
  0.4em\relax USA: Academic Press, Inc., 2010.

\bibitem{Aharon2006_TSP}
M.~Aharon, M.~Elad, and A.~Bruckstein, ``{K-SVD: An algorithm for designing
  overcomplete dictionaries for sparse representation},'' \emph{IEEE
  Transactions on Signal Processing}, vol.~54, no.~11, pp. 4311--4322, 11 2006.

\bibitem{Lee2001_nips}
D.~D. Lee and H.~S. Seung, ``{Algorithms for non-negative matrix
  factorization},'' in \emph{Advances in Neural Information Processing
  Systems}, no.~13, 2001, pp. 556--562.

\bibitem{Ren2017_TAJ}
\BIBentryALTinterwordspacing
B.~R{\'{e}}n, L.~Pueyo, G.~B. Zhu, J.~Debes, and G.~Duch{\^{e}}ne,
  ``{Non-negative matrix factorization: Robust extraction of extended
  structures},'' p. 104, 12 2017. [Online]. Available:
  \url{https://doi.org/10.3847/1538-4357/aaa1f2}
\BIBentrySTDinterwordspacing

\bibitem{Fevotte2009_NR}
C.~F{\'{e}}votte, N.~Bertin, and J.-L. Durrieu, ``{Nonnegative matrix
  factorization with the Itakura-Saito divergence. With application to music
  analysis},'' \emph{Neural Computation}, vol.~21, no.~3, pp. 793--830, 2009.

\bibitem{Taslaman2012_PLOS}
\BIBentryALTinterwordspacing
L.~Taslaman and B.~Nilsson, ``{A Framework for Regularized Non-Negative Matrix
  Factorization, with Application to the Analysis of Gene Expression Data},''
  \emph{PLoS ONE}, vol.~7, no.~11, p. 46331, 11 2012. [Online]. Available:
  \url{https://www.ncbi.nlm.nih.gov/pmc/articles/PMC3487913/}
\BIBentrySTDinterwordspacing

\bibitem{Huang2016_NIPS}
\BIBentryALTinterwordspacing
K.~Huang, X.~Fu, and N.~D. Sidiropoulos, ``{Anchor-Free Correlated Topic
  Modeling: Identifiability and Algorithm},'' in \emph{Advances in Neural
  Information Processing Systems}, D.~Lee, M.~Sugiyama, U.~Luxburg, I.~Guyon,
  and R.~Garnett, Eds., vol.~29.\hskip 1em plus 0.5em minus 0.4em\relax Curran
  Associates, Inc., 2016. [Online]. Available:
  \url{https://proceedings.neurips.cc/paper/2016/file/d707329bece455a462b58ce00d1194c9-Paper.pdf}
\BIBentrySTDinterwordspacing

\bibitem{Fu2019_SPM}
X.~Fu, K.~Huang, N.~D. Sidiropoulos, and W.~K. Ma, ``{Nonnegative Matrix
  Factorization for Signal and Data Analytics: Identifiability, Algorithms, and
  Applications},'' \emph{IEEE Signal Processing Magazine}, vol.~36, no.~2, pp.
  59--80, 2019.

\bibitem{Fevotte2011_NR}
C.~F{\'{e}}votte and J.~Idier, ``{Algorithms for nonnegative matrix
  factorization with the beta-divergence},'' \emph{Neural Computation},
  vol.~23, no.~9, pp. 2421--2456, 2011.

\bibitem{Cichocki2006_LVA}
A.~Cichocki, R.~Zdunek, and S.~Amari, ``{Csiszar's divergences for non-negative
  matrix factorization : Family of new algorithms},'' in \emph{Proc. of
  LVA/ICA}, Charleston, SC, USA, 2006, pp. 32--39.

\bibitem{Wang2010_SIAM}
F.~Wang and P.~Li, ``{Efficient nonnegative matrix factorization with random
  projections},'' in \emph{Proceedings of the 10th SIAM International
  Conference on Data Mining, SDM 2010}.\hskip 1em plus 0.5em minus 0.4em\relax
  Society for Industrial and Applied Mathematics Publications, 2010, pp.
  281--292.

\bibitem{Cemgil2009_CIN}
A.~T. Cemgil, ``{Bayesian Inference for Nonnegative Matrix Factorisation
  Models},'' \emph{Computational Intelligence and Neuroscience}, vol. 2009, p.
  17 pages, 2009.

\bibitem{Hofman2001_ML}
T.~Hofmann, ``{Unsupervised Learning by Probabilistic Latent Semantic
  Analysis},'' \emph{Machine Learning}, vol.~42, pp. 177--196, 2001.

\bibitem{Shashanka2008_CIN}
\BIBentryALTinterwordspacing
M.~Shashanka, B.~Raj, and P.~Smaragdis, ``{Probabilistic Latent Variable Models
  as Nonnegative Factorizations},'' \emph{Computational intelligence and
  neuroscience}, vol. 2008, no.~4, pp. 1--8, 2008. [Online]. Available:
  \url{http://www.ncbi.nlm.nih.gov/pubmed/18509481}
\BIBentrySTDinterwordspacing

\bibitem{Henriet2019_IEEE}
S.~Henriet, U.~Simsekli, S.~D. Santos, B.~Fuentes, and G.~Richard,
  ``{Independent-Variation Matrix Factorization with Application to Energy
  Disaggregation},'' \emph{IEEE Signal Processing Letters}, vol.~26, no.~11,
  pp. 1643--1647, 2019.

\bibitem{Qi2009_BI}
\BIBentryALTinterwordspacing
Q.~Qi, Y.~Zhao, M.~Li, and R.~Simon, ``{Non-negative matrix factorization of
  gene expression profiles: A plug-in for BRB-ArrayTools},''
  \emph{Bioinformatics}, vol.~25, no.~4, pp. 545--547, 2 2009. [Online].
  Available: \url{https://pubmed.ncbi.nlm.nih.gov/19131367/}
\BIBentrySTDinterwordspacing

\bibitem{LeRoux2008_NIPS}
J.~Le~Roux, A.~de~Cheveign{\'{e}}, and L.~C. Parra, ``{Adaptive Template
  Matching with Shift-Invariant Semi-NMF},'' \emph{Advances in Neural
  Information Processing Systems}, vol.~21, 2008.

\bibitem{Trigeorgis2014a_ICML}
\BIBentryALTinterwordspacing
G.~Trigeorgis, K.~Bousmalis, S.~Zafeiriou, and B.~W. Schuller, ``{A Deep
  Semi-NMF model for learning hidden representations},'' in \emph{31st
  International Conference on Machine Learning, ICML 2014}, vol.~5.\hskip 1em
  plus 0.5em minus 0.4em\relax PMLR, 6 2014, pp. 3677--3688. [Online].
  Available: \url{http://proceedings.mlr.press/v32/trigeorgis14.html}
\BIBentrySTDinterwordspacing

\bibitem{Rousset2018_TCI}
F.~Rousset, F.~Peyrin, and N.~Ducros, ``{A Semi Nonnegative Matrix
  Factorization Technique for Pattern Generalization in Single-Pixel
  Imaging},'' \emph{IEEE Transactions on Computational Imaging}, vol.~4, no.~2,
  pp. 284--294, 3 2018.

\bibitem{Ding2010}
\BIBentryALTinterwordspacing
C.~Ding, T.~Li, and M.~Jordan, ``{Convex and semi-nonnegative matrix
  factorizations.}'' \emph{IEEE transactions on pattern analysis and machine
  intelligence}, vol.~32, no.~1, pp. 45--55, 1 2010. [Online]. Available:
  \url{http://www.ncbi.nlm.nih.gov/pubmed/19926898}
\BIBentrySTDinterwordspacing

\bibitem{Chouh2015_LAIA}
\BIBentryALTinterwordspacing
M.~Chouh, M.~Hanafi, and K.~Boukhetala, ``{Semi-nonnegative rank for real
  matrices and its connection to the usual rank},'' \emph{Linear Algebra and
  Its Applications}, vol. 466, pp. 27--37, 2015. [Online]. Available:
  \url{http://dx.doi.org/10.1016/j.laa.2014.09.046}
\BIBentrySTDinterwordspacing

\bibitem{Gillis2015_MAA}
N.~Gillis and A.~Kumar, ``{Exact and heuristic algorithms for semi-nonnegative
  matrix factorization},'' \emph{SIAM Journal on Matrix Analysis and
  Applications}, vol.~36, no.~4, pp. 1404--1424, 2015.

\bibitem{Dreisigmeyer2020_PRIA}
D.~W. Dreisigmeyer, ``{Tight Semi-nonnegative Matrix Factorization},''
  \emph{Pattern Recognition and Image Analysis}, vol.~30, no.~4, pp. 632--637,
  2020.

\bibitem{Blei2003_JMLR}
D.~M. Blei, A.~Y. Ng, and M.~I. Jordan, ``{Latent Dirichlet allocation},''
  \emph{Journal of Machine Learning Research}, vol.~3, no. 4-5, pp. 993--1022,
  2003.

\bibitem{Yilmaz2011_NIPS}
K.~Y. Yilmaz, A.~T. Cemgil, and U.~Simsekli, ``{Generalized Coupled Tensor
  Factorization},'' in \emph{NIPS}, Granada, Spain, 2011.

\bibitem{Mysore2012_ICML}
G.~J.~J. Mysore and M.~Sahani, ``{Variational Inference in Non-negative
  Factorial Hidden Markov Models for Efficient Audio Source Separation},'' in
  \emph{Proc. of ICML}, {\'{E}}dimbourg, {\'{E}}cosse, 2012.

\bibitem{Fuentes2013_TASLP}
B.~Fuentes, R.~Badeau, and G.~Richard, ``{Harmonic Adaptive Latent Component
  Analysis of Audio and Application to Music Transcription},'' \emph{IEEE
  Trans. on Audio, Speech and Language Processing}, vol.~21, no.~9, pp.
  1854--1866, 2013.

\bibitem{Alzaid2010_MMSS}
A.~A. Alzaid and M.~A. Omair, ``{On the Poisson difference distribution
  inference and applications},'' \emph{Bulletin of the Malaysian Mathematical
  Sciences Society}, vol.~33, no.~1, pp. 17--45, 2010.

\bibitem{Gautschi1978_MOC}
W.~Gautschi and J.~Slavik, ``{On the Computation of Modified Bessel Function
  Ratios},'' \emph{Mathematics Of Computation}, vol.~32, no. 143, pp. 865--875,
  1978.

\bibitem{Dempster1977_JRSS}
\BIBentryALTinterwordspacing
A.~P. Dempster, N.~M. Laird, and D.~B. Rubin, ``{Maximum likelihood from
  incomplete data via the EM algorithm},'' \emph{Journal of the Royal
  Statistical Society.}, vol.~39, no.~1, pp. 1--38, 1977. [Online]. Available:
  \url{http://www.jstor.org/stable/10.2307/2984875}
\BIBentrySTDinterwordspacing

\bibitem{Beal2003_PhD}
M.~Beal, ``{Variational Algorithms for Approximate Bayesian Inference},'' Ph.D.
  dissertation, Univ. College of London, 2003.

\bibitem{Cherief-Abdellatif2019_PMLR}
B.~E. Ch{\'{e}}rief-Abdellatif, ``{Consistency of ELBO maximization for model
  selection},'' \emph{Proc. of MLR}, vol.~96, pp. 11--31, 2019.

\bibitem{McGrogy2007_CSDA}
\BIBentryALTinterwordspacing
C.~A. McGrory and D.~M. Titterington, ``{Variational approximations in Bayesian
  model selection for finite mixture distributions},'' \emph{Computational
  Statistics {\&} Data Analysis}, vol.~51, no.~11, pp. 5352--5367, 2007.
  [Online]. Available:
  \url{http://www.sciencedirect.com/science/article/pii/S0167947306002362}
\BIBentrySTDinterwordspacing

\bibitem{Minka2000_TechReport}
\BIBentryALTinterwordspacing
T.~P. Minka, ``{Estimating a Dirichlet distribution},'' Microsoft Research Lab,
  Tech. Rep., 2000. [Online]. Available:
  \url{http://research.microsoft.com/en-us/um/people/minka/papers/dirichlet/minka-dirichlet.pdf}
\BIBentrySTDinterwordspacing

\bibitem{Fuentes2020_web}
\BIBentryALTinterwordspacing
B.~Fuentes, ``{wonterfact (WONderful TEnsoR FACTorization)}.'' [Online].
  Available: \url{https://github.com/SmartImpulse/Wonterfact}
\BIBentrySTDinterwordspacing

\bibitem{Thurau2014_github}
\BIBentryALTinterwordspacing
C.~Thurau, ``{Python Matrix Factorization Module}.'' [Online]. Available:
  \url{https://github.com/cthurau/pymf}
\BIBentrySTDinterwordspacing

\bibitem{Berlinet2012_CSDA}
\BIBentryALTinterwordspacing
A.~F. Berlinet and C.~Roland, ``{Acceleration of the em algorithm: P-EM versus
  epsilon algorithm},'' \emph{Computational Statistics and Data Analysis},
  vol.~56, no.~12, pp. 4122--4137, 2012. [Online]. Available:
  \url{http://dx.doi.org/10.1016/j.csda.2012.03.005}
\BIBentrySTDinterwordspacing

\bibitem{Dua2019_web}
\BIBentryALTinterwordspacing
D.~Dua and C.~Graff, ``{{\{}UCI{\}} Machine Learning Repository},'' 2017.
  [Online]. Available: \url{http://archive.ics.uci.edu/ml}
\BIBentrySTDinterwordspacing

\bibitem{Fitzgerald2009_ISSC}
D.~FitzGerald, M.~Cranitch, and E.~Coyle, ``{On the use of the beta divergence
  for musical source separation},'' in \emph{IET Irish Signals and Systems
  Conference (ISSC 2009)}, 5 2009, pp. 1--6.

\end{thebibliography}

%\begin{IEEEbiographynophoto}{Coauthor}
%Same again for the co-author, but without photo

\end{document}